\def\year{2022}\relax
\documentclass[letterpaper]{article} 
\usepackage{aaai22}  
\usepackage{times}  
\usepackage{helvet}  
\usepackage{courier}  
\usepackage[hyphens]{url}  
\usepackage{graphicx} 
\usepackage{natbib}  
\usepackage{caption} 
\DeclareCaptionStyle{ruled}{labelfont=normalfont,labelsep=colon,strut=off} 
\usepackage{algorithm}
\usepackage{algorithmic}
\usepackage{amsmath}
\usepackage{amssymb}
\usepackage{booktabs}
\usepackage{multirow}
\usepackage{xcolor}
\usepackage{marvosym}
\definecolor{ao(english)}{rgb}{0.0, 0.5, 0.0}
\newcommand{\tableSUPERVISION}{
    \begin{table}[]
        \resizebox{1.0\linewidth}{!}{
            \begin{tabular}{c|l|lllll}
            \toprule
            \multirow{2}{*}{Loss}                & \multirow{2}{*}{Modality} & \multicolumn{5}{c}{IoU with GT (moment or its matched sent.)}\\ \cline{3-7} 
                                                 &                           & 1    & (0.5, 1) & (0, 0.5) & 0    & {\bf other video} \\ \midrule
            \multirow{2}{*}{Mutual Matching} & vid.                      & pos. & -        & neg.     & neg. & neg.        \\
                                                 & {\bf sent.}               & pos. & -        & neg.     & neg. & neg.        \\
                                                 \midrule
            \multicolumn{1}{l|}{IoU Regression}  & vid.                      & max  & high     & low      & min  & -           \\ \bottomrule
            \end{tabular}
        }
        \caption{Supervision signals of our MMN. We categorize moments/sentences as pos./neg. samples according to IoU of this moment (or this sentence's corresponding moment) with the GT moment. We introduce two novel types of supervisions: 1) negative sentence samples in mutual matching scheme (row 2), and 2) inter-video negatives (column 7). }
        \label{tab:supervision}
    \end{table}
}

\newcommand{\tablePairDiscrimination}{
    \begin{table}[t]
        \centering 
        \resizebox{1.0\linewidth}{!}{%
        \scalebox{1.0}{
        \begin{tabular}{c|cc|cc|cc|cc} 
            \toprule
            \multirow{2}{*}{\begin{tabular}[c]{@{}c@{}}BCE\end{tabular}}  &\multicolumn{2}{c}{intra-video}& \multicolumn{2}{c}{inter-video}&   \multicolumn{2}{c}{R@1  }& \multicolumn{2}{c}{R@5  } \\ 
            &vid. & sent. &vid. & sent. &   IoU0.5 & IoU0.7 & IoU0.5 & IoU0.7 \\ 
            \midrule
            \checkmark&& & & & $40.12$ & $23.89$ & $79.57$ &   $53.26$  \\
            
            \checkmark&\checkmark& & &  & $40.65$ & $24.49$ & $77.98$ &   $55.89$ \\ 
            \checkmark&\checkmark& \checkmark& &  & $43.36$ & $26.48$ & $81.85$ &   $57.24$ \\ 
            \checkmark&& &\checkmark & \checkmark& $46.32$ & $26.59$ & $83.41  $ &   $56.77$  \\\hline
            \checkmark&\checkmark&\checkmark & \checkmark&  & $45.13$ & $26.24$ & $\underline{83.69}$ &   $58.13$ \\
            \checkmark&\checkmark&\checkmark & &\checkmark&   $\underline{46.91}$ & $\mathbf{27.46}$ & $83.09$ &   $\underline{58.36}$ \\\hline
            &\checkmark&\checkmark & \checkmark& \checkmark& $44.70$ & $24.92$ & $79.76$ &   $53.71$  \\
            \checkmark&\checkmark&\checkmark & \checkmark&\checkmark&  $\mathbf{47.31}$ & $\underline{27.28}$ & $\mathbf{83.74}$ &   $\mathbf{58.41}$ \\
            \midrule
        \end{tabular}
            }
        }
        
        \resizebox{1.0\linewidth}{!}{%
        \scalebox{1.0}{
        \begin{tabular}{c|cc|cc|ccc|ccc} 
            \toprule
            \multirow{2}{*}{\begin{tabular}[c]{@{}c@{}}BCE\end{tabular}} &\multicolumn{2}{c}{intra-video}& \multicolumn{2}{c}{inter-video}&   \multicolumn{3}{c}{R@1  }& \multicolumn{3}{c}{R@5  } \\ 
            &vid. & sent. &vid. & sent. &  IoU0.3 & IoU0.5 & IoU0.7 & IoU0.3 &IoU0.5 & IoU0.7 \\ 
            \midrule
            \checkmark&& & & & $62.73$ & $46.74$ & $27.12$ &   $85.67$ & $77.26$ & $61.64$\\
            \checkmark&\checkmark& & &  & $62.27$ & $46.43$ & $27.75$ & $85.80$ & $77.63$& $61.68$\\ 
            \checkmark&\checkmark& \checkmark& &  & $63.41$ & $47.52$ & $\mathbf{29.69}$ &   $86.37$ & $78.17$& $62.50$\\
            \checkmark&& &\checkmark& \checkmark  &$\underline{64.64}$ & $\underline{48.26}$ & $28.75$ &   $86.52$ &   $78.82$ &   $\underline{64.49}$\\\hline
            \checkmark&\checkmark&\checkmark & \checkmark&  & $64.01$ & $47.47$ & $28.89$ &   $\mathbf{87.38}$ & $\underline{79.26}$ &$64.05$\\
            \checkmark&\checkmark&\checkmark & &\checkmark&  $64.25$ & $48.24$ & $28.96$ & $86.65$ & $78.17$& $62.62$\\\hline
            &\checkmark& \checkmark&\checkmark& \checkmark  &$63.47$ & $45.28 $ & $26.29$ &   $85.55 $ &   $76.89$ &   $60.62$\\
            \checkmark&\checkmark&\checkmark & \checkmark&\checkmark&  $\mathbf{65.05}$ & $\mathbf{48.59}$ & $\underline{29.26}$ &   $\underline{87.25}$&   $\mathbf{79.50}$&   $\mathbf{64.76}$\\
            \bottomrule
        \end{tabular}
        }
    }
    \caption{Ablation on negative samples in our mutual matching scheme. (top) Charades-STA; (bottom) ActivityNet Captions. Intra/inter-video means we use video moments (vid.) or sentences (sent.) inside or cross videos as negatives.}
    \label{tab:pd}
    \end{table}
}
\newcommand{\tableNUMBER}{
\begin{table}[t]
\hspace{-0.1cm}
    \centering 
    \resizebox{1.05\linewidth}{!}{
        \scalebox{1.0}{
            \begin{tabular}{l|ccc|ccc} 
            \toprule
            \multirow{2}{*}{{ Method}}&   \multicolumn{3}{c|}{R@1  }& \multicolumn{3}{c}{R@5  } \\ 
            &   IoU0.3 &IoU0.5 & IoU0.7 & IoU0.3 &IoU0.5 & IoU0.7 \\ 
            \midrule
            Only Intra&$63.41$ & $47.52$ & $\mathbf{29.69}$ &   $86.37$ & $78.17$& $62.50$\\
            \#neg. = Intra & $\underline{64.18}$ & $\underline{47.55}$ & $28.48$ &   $\underline{86.74}$ &   $\underline{79.01}$ &   $\underline{64.15}$\\
            Intra + Inter  & $\mathbf{65.05}$ & $\mathbf{48.59}$ & $\underline{29.26}$ &   $\mathbf{87.25}$ &   $\mathbf{79.50}$ &   $\mathbf{64.76}$\\
            
            \bottomrule
            \end{tabular}
        }
    }
\caption{Ablation on number of negatives (ActivityNet).}
\label{tab:number}
\end{table}
}

\newcommand{\tableAGGREGATION}{
    \begin{table}[t]
        \centering 
        \resizebox{0.8\linewidth}{!}{%
        \scalebox{1.0}{
        \begin{tabular}{l|cc|cc} 
        \toprule
        \multirow{2}{*}{Charades-STA}&   \multicolumn{2}{c|}{R@1  }& \multicolumn{2}{c}{R@5  } \\ 
         &   IoU0.5 & IoU0.7 & IoU0.5 & IoU0.7 \\ 
        \midrule
        Class Embedding `CLS' & $45.11$ & $27.25$ & $82.37$ &   $58.09$ \\
        Average Pooling & $\mathbf{47.31}$ & $\mathbf{27.28}$ & $\mathbf{83.74}$ &   $\mathbf{58.41}$ \\
        \bottomrule
        \toprule
        $m=0.3$ & $45.30$ & $27.13$ & $83.33$ &   $\mathbf{59.02}$ \\
        $m=0.4$ & $\mathbf{47.31}$ & $\mathbf{27.28}$ & $\mathbf{83.74}$ &   $58.41$ \\
        $m=0.5$ & $46.32$ & $26.82$ & $83.09$ &   $58.10$ \\
        \midrule
        $\lambda = 0.05$ &$\mathbf{47.31}$ & $\mathbf{27.28}$ & $\mathbf{83.74}$ &   $\mathbf{58.41}$ \\
        $\lambda = 0.1$ & $45.63$ & $27.13$ & $83.34$ &   $57.33$ \\
        $\lambda = 0.2$ & $44.91$ & $26.95$ & $82.97$ &   $57.69$ \\
        \midrule
        $\tau_v=\tau_s=0.1$ & $\mathbf{47.31}$ & $\mathbf{27.28}$ & $\mathbf{83.74}$ &   $\mathbf{58.41}$ \\
        $\tau_v=\tau_s=0.2$ & $43.88$ & $25.81$ & $82.23$ &   $57.55$ \\
        \bottomrule
        \end{tabular}
        }
    }
    \caption{Ablation on (top) text feature aggregations and (bottom) hyper-parameters in the loss function.}
    \label{tab:aggregation}
    \end{table}
}

\newcommand{\tableANET}{
    \begin{table}[t]
        \centering 
        \resizebox{1\linewidth}{!}{%
        \scalebox{1.0}{
        \begin{tabular}{l|ccc|ccc} 
        \toprule
        \multirow{ 2}{*}{Method}&   \multicolumn{3}{c|}{R@1  }& \multicolumn{3}{c}{R@5  } \\ 
        &   IoU0.3 &IoU0.5 & IoU0.7 & IoU0.3 &IoU0.5 & IoU0.7 \\ 
        \midrule
        MCN \cite{DBLP:conf/iccv/HendricksWSSDR17} & $39.35$ &$21.36$ & $6.43$ & $68.12$ &  $53.23$ & $29.70$  \\
        CTRL \cite{DBLP:conf/iccv/GaoSYN17}  & $47.43$&$29.01$ & $10.34$ & $75.32$ & $59.17$ & $37.54$  \\
        QSPN \cite{DBLP:conf/aaai/Xu0PSSS19}  & $52.13$&$33.26$ & $13.43$ & $77.72$& $62.39$ & $40.78$  \\
        SCDM~\cite{DBLP:conf/nips/YuanMWL019} & $54.80$&$36.75$ & $19.86$ & $77.29$ & $64.99$ & $41.53$ \\
        PMI \cite{DBLP:conf/eccv/ChenJLJ20} & $59.69$&$38.28$ & $17.83$ & $-$&$-$ & $-$  \\
        LGI~\cite{DBLP:conf/cvpr/MunCH20} & $58.52$&$41.51$ & $23.07$ & $-$&$-$ & $-$  \\
        CMIN \cite{DBLP:journals/tip/LinZZZC20}  & $63.61$&$44.62$ & $24.48$ &$82.39$& $69.66$ & $52.96$  \\
        DRN~\cite{DBLP:conf/cvpr/ZengXHCTG20} & $-$&$45.45$ & $24.39$ & $-$ & $77.97$ & $50.30$ \\
        2D-TAN~\cite{DBLP:conf/aaai/ZhangPFL20} & $59.46$&$44.51$ & $26.54$ & $85.53$& $77.13$ &  $61.96$ \\
        IVG-DCL~\cite{DBLP:conf/cvpr/NanQXLLZL21}& $63.22$ &$43.84$& $27.10$ &$-$ &$-$ &$-$\\
        CBLN~\cite{DBLP:conf/cvpr/LiuQDZ00XX21} & $\mathbf{66.34}$ &$\underline{48.12}$ &$27.60$& $\mathbf{88.91}$ &$\underline{79.32}$ &$\underline{63.41}$\\
        CPN~\cite{DBLP:conf/cvpr/ZhaoZZL21} &$62.81$ &$45.10$ &$\underline{28.10}$ &$-$ &$-$ &$-$\\
        FVMR~\cite{Gao_2021_ICCV} &$60.63$ &$45.00$ &$26.85$ &$86.11$ &$77.42$ &$61.04$\\
        SSCS~\cite{Ding_2021_ICCV}&$61.35$ &$46.67$ &$27.56$ &$86.89$ &$78.37$ &$63.78$\\
        \midrule
        Our MMN  & $\underline{65.05}$ & $\mathbf{48.59}$ & $\mathbf{29.26}$ &   $\underline{87.25}$ &   $\mathbf{79.50}$ &   $\mathbf{64.76}$\\
        \bottomrule
        \end{tabular}
        }
    }
    \caption{Performance comparison on ActivityNet Captions (C3D feature).}
    \label{tab:anet}
    \end{table}
}

\newcommand{\tableCHARADES}{
    \begin{table}[t]
        \centering 
        \resizebox{1.0\linewidth}{!}{%
        \scalebox{1.0}{
        \begin{tabular}{l|cc|cc} 
        \toprule
        \multirow{2}{*}{Method} &  \multicolumn{2}{c|}{R@1  }& \multicolumn{2}{c}{R@5  } \\ 
          &IoU0.5 & IoU0.7 & IoU0.5 & IoU0.7 \\ 
        \midrule
        MCN~\cite{DBLP:conf/iccv/HendricksWSSDR17} &  $17.46$ & $8.01$ &  $48.22$ & $26.73$  \\
        
        SAP~\cite{DBLP:conf/aaai/ChenJ19a} &  $27.42$ & $13.36$ &  $66.37$ & $38.15$  \\
        MAN~\cite{DBLP:conf/cvpr/ZhangDWWD19} &  $41.21$ & $20.54$ &  $83.21$ & $51.85$  \\
        {\color{gray} DRN~\cite{DBLP:conf/cvpr/ZengXHCTG20}}$^\dagger$ & ${\color{gray}42.90}$ & ${\color{gray}23.68}$ & ${\color{gray}\underline{87.80}}$ & ${\color{gray}54.87}$ \\
        2D-TAN~\cite{DBLP:conf/aaai/ZhangPFL20} & $39.70$ & $23.31$ & $80.32$ &   $51.26$ \\ 
        CBLN~\cite{DBLP:conf/cvpr/LiuQDZ00XX21} &   $43.67$& $24.44$ &$\mathbf{88.39}$ &$\underline{56.49}$\\
        CPN~\cite{DBLP:conf/cvpr/ZhaoZZL21} &   $\underline{46.08}$& $25.06$& $-$& $-$\\
        FVMR~\cite{Gao_2021_ICCV} &   $42.36$& $24.14$ &$83.97$ &$50.15$\\
        SSCS~\cite{Ding_2021_ICCV} &   $43.15$& $\underline{25.54}$& $84.26$& $54.17$\\   
        \midrule
        Our MMN  & $\mathbf{47.31}$ & $\mathbf{27.28}$ & $83.74$ &   $\mathbf{58.41}$ \\
        \bottomrule
        \end{tabular}
        }
        }
    \caption{Performance comparison on Charades-STA (VGG feature). $^\dagger$our re-production with official code is much lower than the value reported in paper.}
    \label{tab:charades}
    \end{table}
}

\newcommand{\tableTACOS}{
    \begin{table}[t]
        \centering 
        \resizebox{\linewidth}{!}{%
        \scalebox{0.9}{
        \begin{tabular}{l|ccc|ccc} 
        \toprule
        \multirow{2}{*}{Method}&   \multicolumn{3}{c|}{R@1  }& \multicolumn{3}{c}{R@5  } \\ 
        &  IoU0.1 & IoU0.3 & IoU0.5 & IoU0.1 & IoU0.3 & IoU0.5 \\ 
        \midrule
        MCN \cite{DBLP:conf/iccv/HendricksWSSDR17} & $14.42$ & $-$ & $5.58$ & $37.35$ & $-$ & $10.33$  \\
        CTRL \cite{DBLP:conf/iccv/GaoYSCN17}  & $24.32$ & $18.32$ & $13.30$ & $48.73$ & $ 36.69$ & $25.42$  \\
        QSPN \cite{DBLP:conf/aaai/Xu0PSSS19}  & $25.31$ & $20.15$ & $15.23$ & $53.21$ & $36.72$ & $25.30$  \\
        SCDM~\cite{DBLP:conf/nips/YuanMWL019} &$-$ & $26.11$ & $21.17$ & $-$ & $43.35$ & $28.53$  \\
        CMIN~\cite{DBLP:journals/tip/LinZZZC20}  & $36.68$ & $27.33$ & $19.57$ & $64.93$ & $40.16$ & $32.18$  \\
        DRN~\cite{DBLP:conf/cvpr/ZengXHCTG20} &  $-$ & $-$ & $23.17$ & $-$ & $-$ & $33.36$ \\
        2D-TAN~\cite{DBLP:conf/aaai/ZhangPFL20} &  $47.59$ & $37.29$ & $25.32$ & $70.31$ & $57.81$ &  $45.04$ \\
        IVG-DCL~\cite{DBLP:conf/cvpr/NanQXLLZL21}  &$49.36$ &$38.84$ &$29.07$ &$-$ &$-$& $-$ \\
        CBLN~\cite{DBLP:conf/cvpr/LiuQDZ00XX21} &$49.16$ &$38.98$ &$27.65$ &$73.12$& $59.96$ &$46.24$\\
        FVMR~\cite{Gao_2021_ICCV} & $\mathbf{53.12}$& $\mathbf{41.48}$& $\underline{29.12}$& $\mathbf{78.12}$& $\mathbf{64.53}$& $\mathbf{50.00}$\\
        SSCS~\cite{Ding_2021_ICCV} &   $50.78$& $\underline{41.33}$& $\mathbf{29.56}$& $72.53$& $60.65$& $\underline{48.01}$\\
        \midrule
        Our MMN  &  $\underline{51.39}$ & $39.24$ & $26.17$ & $\underline{78.03}$ & $\underline{62.03}$ & $47.39$ \\
        \bottomrule
        \end{tabular}
        }
        }
        \caption{Performance comparison on TACoS (C3D feature).}
        \label{tab:tacos}
    \end{table}
}
\newcommand{\tableSTVG}{
\begin{table}[t]
    \begin{center}
    \resizebox{1.0\linewidth}{!}{%
            \begin{tabular}{l|c|c|c}
            \toprule
            Method & m\_vIoU & vIoU@0.3 & vIoU@0.5 \\
            \midrule
            2D-TAN + WSSTG$^\dagger$ & $15.43$ & $19.83$ & $6.81$ \\
            STGVT~\cite{DBLP:journals/corr/abs-2011-05049}$^\dagger$ & $18.15$& $26.81$ & $9.48$ \\
            STVGBert~\cite{Su_2021_ICCV} & $20.42$ & $29.37$ & $11.31$ \\
            Yu \etal~\cite{DBLP:journals/corr/abs-2106-07166} & $\underline{30.02}$& $-$&$-$\\
            {\color{gray}Tan \etal ~\cite{DBLP:journals/corr/abs-2106-10634}}$^\ddagger$ & {\color{gray}$30.40$}& {\color{gray}$50.40$}&{\color{gray}$18.80$} \\
            Our stage-1+2D-TAN & $22.83$ & $\underline{36.07}$ & $\underline{16.96}$ \\
            \midrule
            Our stage-1+MMN & $\mathbf{30.32}$ & $\mathbf{49.02}$ & $\mathbf{25.56}$\\
            \bottomrule
            \end{tabular}
    }
    \end{center}
    \caption{Performance comparison on HC-STVG val set. $^\dagger$use smaller train set, values reported in~\cite{DBLP:journals/corr/abs-2011-05049}. $^\ddagger$uses an ensemble of 10 models.}
    \label{table:stvg}
\end{table}
}

\newcommand{\tableDistilBERT}{
    \begin{table}[t]
        \centering 
        \resizebox{1.0\linewidth}{!}{%
        \scalebox{0.9}{
        \begin{tabular}{l|c|cc|cc} 
        \toprule
        \multirow{2}{*}{Charades-STA}& \multirow{2}{*}{Feature} & \multicolumn{2}{c|}{R@1  }& \multicolumn{2}{c}{R@5  } \\ 
         &  & IoU0.5 & IoU0.7 & IoU0.5 & IoU0.7 \\ 
        \midrule
        DRN$^\dagger$~\cite{DBLP:conf/cvpr/ZengXHCTG20} &C3D& $\mathit{43.26}$ & $\mathit{23.95}$ & $85.43$ &   $52.82$ \\
        DRN w/ DistilBERT &C3D& $42.62$ & $23.50$ & $\mathit{85.84}$ &   $\mathit{53.59}$ \\
        \midrule
        \midrule
        2D-TAN \cite{DBLP:conf/aaai/ZhangPFL20} &VGG& $39.70$ & $23.31$ & $80.32$ &   $51.26$ \\
        2D-TAN$^\dagger$ &VGG& $40.03$ & $23.15$ & $\underline{82.18}$ &   $49.89$ \\ 
        2D-TAN w/ DistilBERT &VGG& $\underline{40.32}$ & $\underline{23.81}$ & $79.49$ &   $\underline{51.96}$ \\
        \bottomrule
        \end{tabular}
        }
    }
    \caption{Previous methods with DistilBERT. $^\dagger$ means our implementation.}
    \label{tab:bert}
    \end{table}
}

\newcommand{\figFIRST}{
    \begin{figure}[t]
        \begin{center}
        \includegraphics[width=8cm]{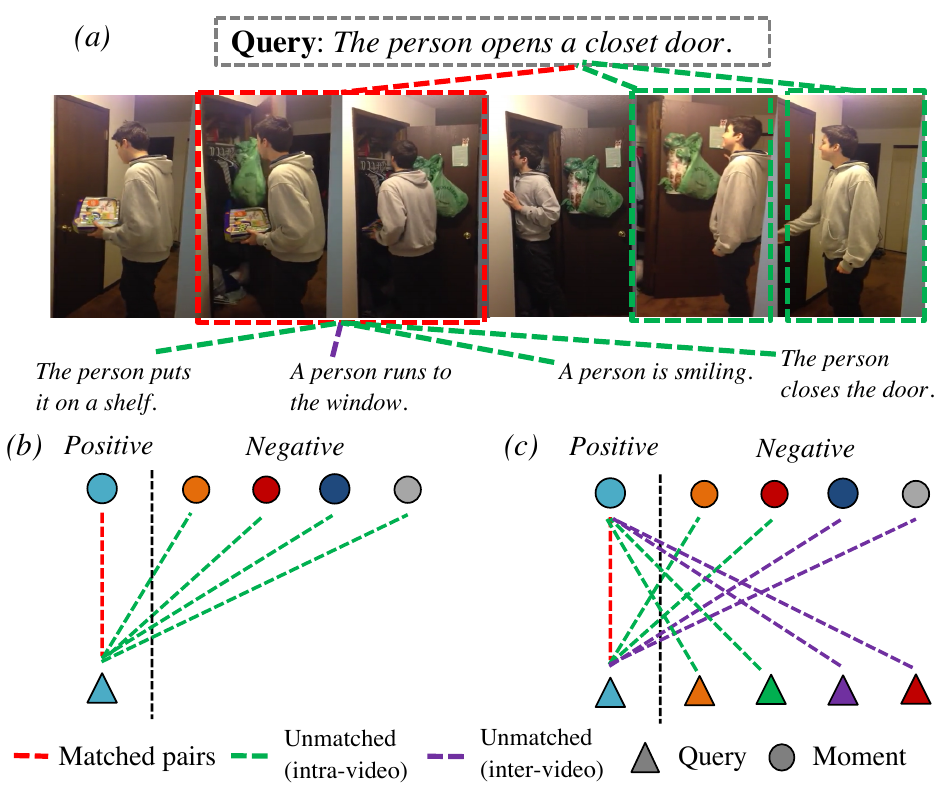}
        \end{center}
        \caption{(a) In addition to matching moments given a query, we propose a new auxiliary task named {\em mutual matching} to differentiate the matched query (red dash line) among unmatched queries intra- (green) or inter-video (purple) for the GT moment (red box). (b) Most previous methods only consider intra-video negative moments (green). (c) Our {\em mutual matching} uses negative pairs from both modalities intra- (green) or inter-video (purple). {\bf Best view with colors}.}
        \label{fig:first}
    \end{figure}
        
}

\newcommand{\figOVERVIEW}{
    \begin{figure*}[t]
        \begin{center}
        \includegraphics[width=1.0\linewidth]{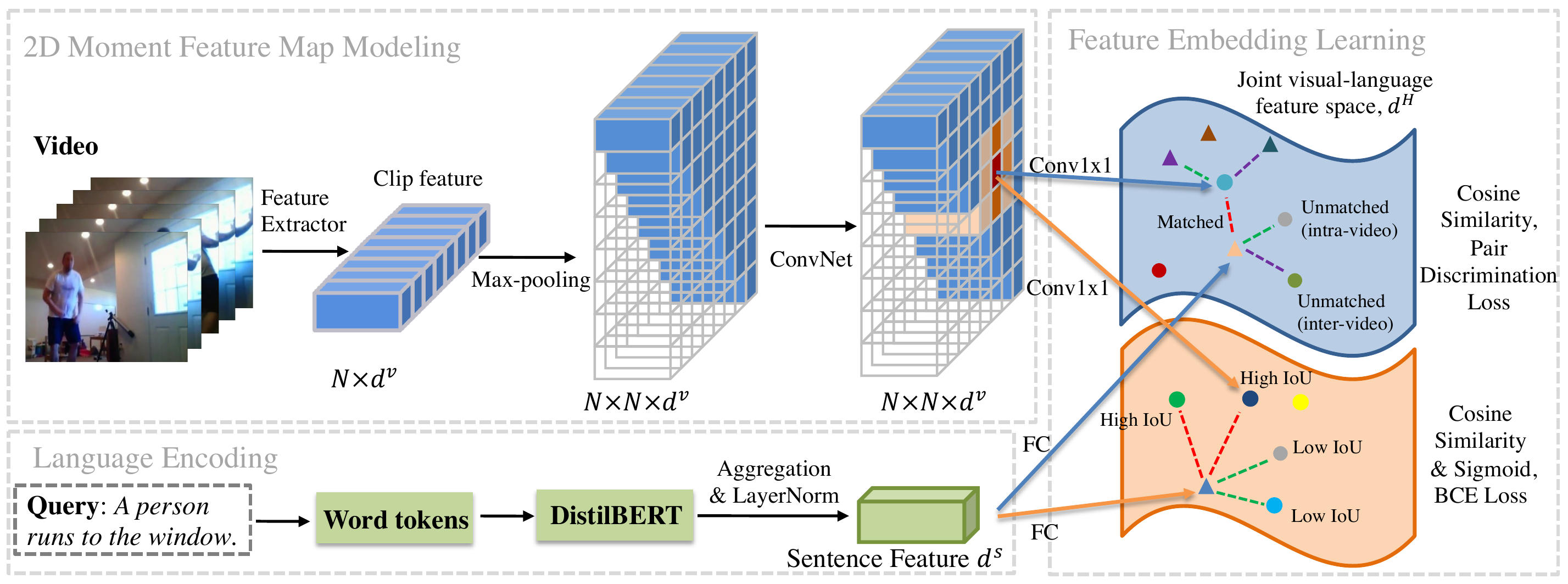}
        \end{center}
           \caption{Overview of our framework. Different from the baseline, we adopt a late modality fusion strategy and learning the feature embedding for moments and sentences on two independent spaces with pair discrimination and BCE losses, respectively. The dots and triangles are the feature of moments and sentences. The red dash lines are matched moment-sentence pairs to be pulled in, while green/purple dash lines are negative samples intra/inter-video to be pushed away. {\bf Best view with colors}.}
        \label{fig:overview}
    \end{figure*}
}

\newcommand{\figSANITY}{
    \begin{figure}[t]
            \begin{center}
            \includegraphics[height=3cm]{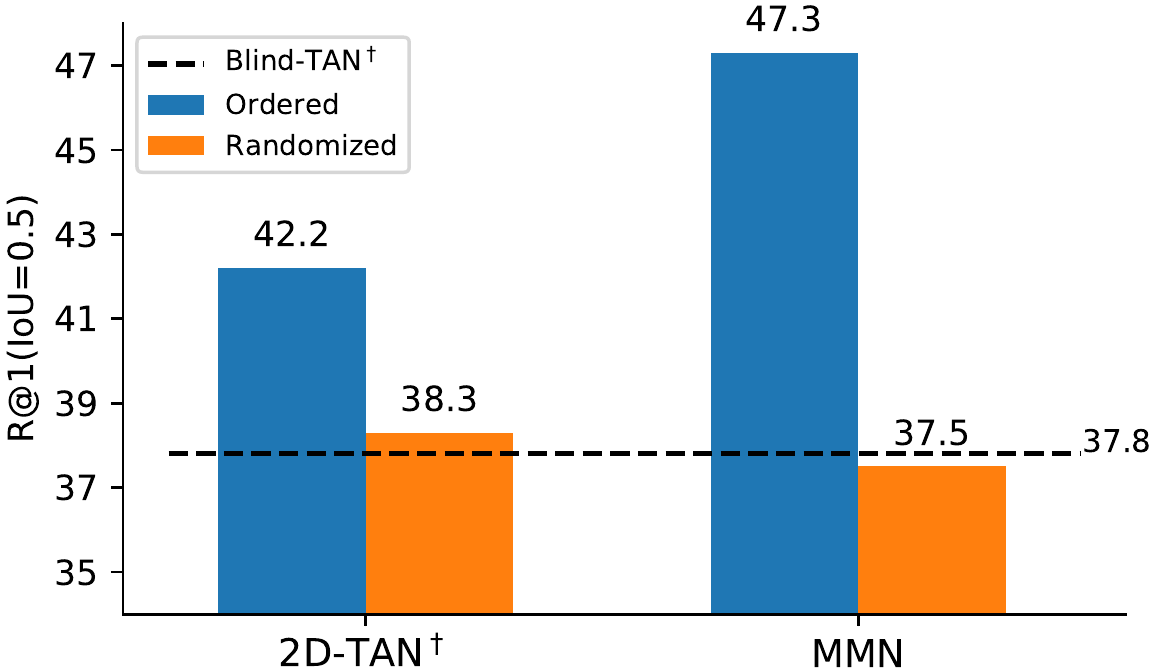}
            \end{center}
            \caption{Sanity Check on the Charades-STA dataset. $^\dagger$values are obtained from~\cite{DBLP:conf/bmvc/OtaniNRH20}.}
            \label{fig:sanity}
    \end{figure}
}

\newcommand{\figVISUAL}{
    \begin{figure*}[ht]
        \begin{center}
        \includegraphics[width=17cm]{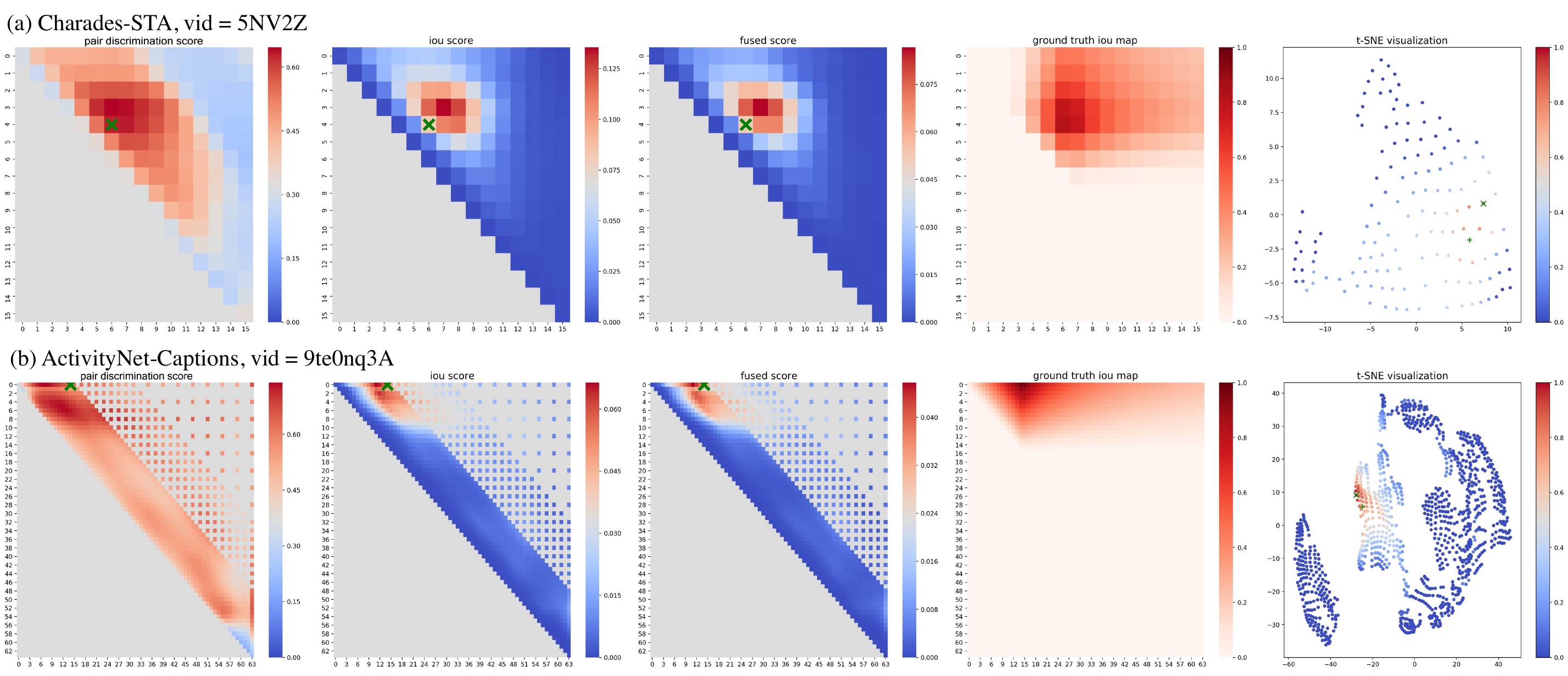}
        \end{center}
        \caption{Visualizations on two datasets. Green `{\color{ao(english)}x}' in the first three maps shows location of GT moment. In the last map, the matched moment (i.e., has the highest IoU among candidate moments) is a green `{\color{ao(english)}x}' and the given sentence is a green `{\color{ao(english)}+}'. The colors of unmatched moments shows their IoU with GT moment.}
        \label{fig:visual_anet_charades}
    \end{figure*}
}

\newcommand{\etal}{\textit{et al.}}
\usepackage{newfloat}
\usepackage{listings}
\floatstyle{ruled}
\newfloat{listing}{tb}{lst}{}
\floatname{listing}{Listing}
\title{Negative Sample Matters: A Renaissance of Metric Learning \\ for Temporal Grounding}
\author{Zhenzhi Wang \quad Limin Wang\thanks{Corresponding author.}\quad Tao Wu\quad  Tianhao Li\quad Gangshan Wu}
\affiliations{State Key Laboratory for Novel Software Technology, Nanjing University, China \\
{\tt\small \{zhenzhiwang,tianhaolee\}@outlook.com, \{lmwang,gswu\}@nju.edu.cn, wt@smail.nju.edu.cn}
}
\begin{document}
\maketitle

\begin{abstract}
Temporal grounding aims to localize a video moment which is semantically aligned with a given natural language query. Existing methods typically apply a detection or regression pipeline on the fused representation with the research focus on designing complicated prediction heads or fusion strategies. Instead, from a perspective on temporal grounding as a metric-learning problem, we present a Mutual Matching Network (MMN), to directly model the similarity between language queries and video moments in a joint embedding space. This new metric-learning framework enables fully exploiting negative samples from two new aspects: constructing negative cross-modal pairs in a mutual matching scheme and mining negative pairs across different videos. These new negative samples could enhance the joint representation learning of two modalities via cross-modal mutual matching to maximize their mutual information. Experiments show that our MMN achieves highly competitive performance compared with the state-of-the-art methods on four video grounding benchmarks. Based on MMN, we present a winner solution for the HC-STVG challenge of the 3rd PIC workshop. This suggests that metric learning is still a promising method for temporal grounding via capturing the essential cross-modal correlation in a joint embedding space. Code is available at \url{https://github.com/MCG-NJU/MMN}.
\end{abstract}

\section{Introduction}
\label{sec:intro}
Video analysis is a fundamental problem in computer vision and has drawn increasing attention in recent years because of the potential applications in surveillance, robotics, and Internet videos. While considerable progress has been made in video classification~\cite{DBLP:conf/eccv/WangXW0LTG16,DBLP:conf/cvpr/WangL0G18,DBLP:conf/iccv/Feichtenhofer0M19,DBLP:conf/cvpr/0002TJW21} and action localization~\cite{DBLP:conf/iccv/ZhaoXWWTL17,DBLP:conf/iccv/LinLLDW19,DBLP:conf/eccv/LiW0W20}, such tasks are still limited to recognizing a pre-defined list of activities, such as playing baseball or peeling potatoes. As videos often contain complex activities that may cause a combinatorial explosion if described by a list of actions and objects, the task of grounding language queries in videos~\cite{DBLP:conf/iccv/HendricksWSSDR17,DBLP:conf/iccv/GaoSYN17,DBLP:journals/corr/abs-2011-05049}, as a generalization of the action localization task to overcome the constraints, has recently gained plenty of interest~\cite{DBLP:conf/aaai/HeZHLLW19,DBLP:conf/cvpr/ZhangDWWD19,DBLP:conf/cvpr/ZengXHCTG20,DBLP:conf/nips/YuanMWL019,DBLP:conf/aaai/ZhangPFL20,DBLP:journals/corr/abs-2011-05049} in both computer vision and language community. Formally, given a verbal description, the goal of temporal grounding is to determine the temporal moment (i.e., the start and end time) that semantically corresponds to the query best in a given video.

\figFIRST
Although temporal grounding opens up great opportunities for detailed video perception by using the new language modality to capture the complex relations between sentences and videos, most of the previous approaches~\cite{DBLP:conf/cvpr/ZhangDWWD19,DBLP:conf/nips/YuanMWL019,DBLP:conf/aaai/ZhangPFL20,DBLP:conf/cvpr/ZengXHCTG20,DBLP:journals/corr/abs-2011-05049} still tackle this problem in a {\em detection/regression} way with {\em early-fusion} designs, e.g., using the fused multi-modal features to predict the offset of action moments from anchors; or directly regressing the desired region on a globally aggregated multi-modal feature. These indirect grounding methods typically ignore the essential relation between all cross-modal pairs (i.e., pairs of moment and language queries), and they only simply utilize the IoU-based scores between the given sentence and moments from the {\em same video} as supervision (Fig.~\ref{fig:first}(b)). However, we argue that the negative relations between the given moment and other {\em unmatched descriptions} are also important for learning a joint cross-modal embedding. Intuitively, the auxiliary task, given a video moment with visually informative actions, training the model to contrast the matched query and unmatched descriptions, (Fig.~\ref{fig:first}(a)) is beneficial for the temporal grounding task. We propose to use this auxiliary task for the first time in temporal grounding, which shows advantages in less extra computational cost compared with previous auxiliary tasks. In order to model such relations from mutual directions (Fig.~\ref{fig:first}(c)), we revisit the temporal grounding task from a {\em metric-learning perspective}, which allows us to directly model the essential similarity measurement in joint cross-modal embedding space, rather than designing sophisticated detection/regression strategies based on the fused representation as in many previous methods.  

In our metric-learning view, sentences and moments play an equally important role in temporal grounding and both matching directions are adopted, i.e., select the right instance in a modality given the groundtruth of another modality. Thus, supervisions are constructed in a symmetric form for two modalities by our approach (Fig.~\ref{fig:first}(c)), creating more supervisions than previous methods. Our framework shows several advantages: Firstly, the metric-learning perspective enables us to mine negative samples from a {\em mutual matching scheme}. In this sense, an {\em unmatched relation} is also informative by implying that this moment and sentence should be pushed away in the joint space. To achieve this, we adopt a cross-modal mutual matching objective to contrast pos./neg. moment-sentence pairs inspired by a cross-modal pretraining method~\cite{DBLP:journals/corr/abs-2001-05691}. 
Secondly, we exploit much more negative samples to improve our representation learning by adopting {\em inter-video} negative samples. In contrast, most of the previous methods only utilize {\em intra-video} moment-sentence pairs as supervision due to the early-fusion strategy.
Finally, we model the video and language feature in a Siamese-alike network architecture and use a simple dot-product for cross-modal similarity computation. Thus, our framework has less computation cost by sharing moment features among sentences inside each video.

While using the negative pairs can lead to a better feature, the binary supervision signal itself (i.e., matched or unmatched) is still weak to precisely rank the moments inside a video. In addition, although the negative pairs used in our mutual matching scheme provide much more supervision signals, most of them are easy negatives (especially the inter-video ones). So we adopt another embedding space for learning the cross-modal similarity for precisely ranking moments based on the IoU supervision, which enables us to estimate finer relations (i.e., a scalar instead of a binary signal) and accurately differentiate the correct pair among hard negatives (e.g., negative moments with high IoU). These two complementary objectives share the same feature encoders as backbone and have their own customized heads.

Our main contributions are three-fold: (1) We revisit the metric-learning perspective on the temporal grounding task in a {\em late-fusion} fashion and leverage {\bf inter-video} negative visual-language pairs. Our framework shows advantages on both performance and training cost. (2) Our metric-learning view enables a new auxiliary task for temporal grounding, named {\bf cross-modal mutual matching} to significantly add more supervision signals. Instead of heavy additional networks for previous auxiliary tasks, ours has a more direct idea and is also more effective. (3) Comprehensive quantitative and qualitative analyses are conducted on four benchmarks from both temporal and spatio-temporal video grounding tasks to show our method's generalization ability.

\section{Related Works}
\subsection{Temporal Grounding}
\noindent\textbf{Temporal Grounding Methods}. Previous methods can be mainly categorized into four groups: \textbf{(1) Regression based methods} directly predict the boundaries (i.e., start and end) of the target moment from the fused multi-modal features relying on either a clip-wise boundary classification in local regions~\cite{DBLP:conf/naacl/GhoshAPH19,DBLP:conf/wacv/OpazoMSLG20, DBLP:conf/acl/ZhangSJZ20,DBLP:conf/aaai/ChenLTXZTL20} or a direct boundary regression on the aggregated global feature~\cite{DBLP:conf/aaai/YuanM019,DBLP:conf/aaai/Wang0J20,DBLP:conf/cvpr/MunCH20}. Some methods also introduce some heuristics, e.g., aggregating clip-wise actionness scores modeled by compositional reasoning~\cite{DBLP:conf/eccv/LiuYCHFN18} or taking the expectation of probability for start/end~\cite{DBLP:conf/naacl/GhoshAPH19}. \textbf{(2) Detection based methods} often first generate candidate moments and then evaluate them on the fused multi-modal features~\cite{DBLP:conf/iccv/GaoSYN17, DBLP:conf/wacv/GeGCN19}. Their evaluation on moments utilize various designs, e.g., LSTM~\cite{DBLP:conf/emnlp/ChenCMJC18,DBLP:conf/aaai/Xu0PSSS19}, dynamic filtering~\cite{DBLP:conf/cvpr/ZhangDWWD19,DBLP:conf/wacv/OpazoMSLG20} or modulation~\cite{DBLP:conf/nips/YuanMWL019}, graph convolution~\cite{DBLP:conf/cvpr/ZhangDWWD19}, anchor-free detectors~\cite{DBLP:conf/emnlp/LuCTLX19,DBLP:conf/cvpr/ZengXHCTG20}, and 2D moment map~\cite{DBLP:conf/aaai/ZhangPFL20}.  \textbf{(3) Reinforcement Learning based methods}~\cite{DBLP:conf/aaai/HeZHLLW19,DBLP:conf/cvpr/WangHW19,DBLP:conf/aaai/WuLLL20} localize the target moment iteratively by defining the {\em state} and {\em action} on the video and treating this task as a sequential decision-making process. \textbf{(4)} The only previous \textbf{metric learning based method}~\cite{DBLP:conf/iccv/HendricksWSSDR17} uses a triplet loss with $\ell_2$ distance as the similarity measurement to match the given sentence to the correct video moment. However, it lacks both the important supervision of negative sentence samples, i.e., our novel {\em mutual matching scheme} and the effective relation modeling of moments, thus achieves much worse results than ours.

\noindent\textbf{Multi-modal Fusion.} Many methods from the group 1) to 3) mainly adopt a {\em early-fusion} pipeline for cross-modal modeling, e.g., by concatenation~\cite{DBLP:conf/emnlp/ChenCMJC18,DBLP:journals/tip/LinZZZC20,DBLP:conf/aaai/Wang0J20}, dynamic convolution~\cite{DBLP:conf/wacv/OpazoMSLG20,DBLP:conf/cvpr/ZhangDWWD19,DBLP:conf/nips/YuanMWL019}, cross-attention~\cite{DBLP:conf/emnlp/LuCTLX19,DBLP:journals/corr/abs-2009-11232}, or hadamard product~\cite{DBLP:conf/cvpr/ZengXHCTG20,DBLP:conf/cvpr/MunCH20,DBLP:conf/aaai/ZhangPFL20}. On the contrary, we use a simple inner-product in the joint visual-language space to measure the cross-modal similarity in a {\em late-fusion} manner. It not only enables our mutual matching scheme, but also shows advantages in computational cost during training by sharing the video feature among sentences in the same video.

\noindent\textbf{Auxiliary Tasks in Temporal Grounding.} Query reconstruction as an auxiliary task for temporal grounding was explored by~\cite{DBLP:journals/tip/LinZZZC20,DBLP:conf/aaai/Xu0PSSS19} who added a video caption loss following image-based grounding methods~\cite{DBLP:conf/cvpr/RamanishkaDZS17}. Yet it introduces extra overhead of parameters and computational cost, e.g., LSTM~\cite{DBLP:conf/aaai/Xu0PSSS19} or Transformer~\cite{DBLP:journals/tip/LinZZZC20}. Our proposed auxiliary task of mutual matching avoid the usage of the heavy additional network and is also more direct and effective. 

\subsection{Spatio-Temporal Video Grounding}
As the recent progress achieved in spatio-temporal action localization~\cite{DBLP:conf/eccv/LiW0W20}, spatio-temporal video grounding~\cite{DBLP:journals/corr/abs-2011-05049} is also proposed as an extension of temporal grounding. By adapting our MMN on the linked human bounding boxes~\cite{DBLP:conf/iccv/KalogeitonWFS17a} for temporal trimming, significant performance gain is achieved over the previous transformer-based method~\cite{DBLP:journals/corr/abs-2011-05049}.

\subsection{Metric Learning}
The family of metric learning loss has been explored to learn powerful representations with the supervised setting~\cite{DBLP:conf/cvpr/HadsellCL06,DBLP:conf/nips/KhoslaTWSTIMLK20}, where the positive sample is chosen from the same class and the negative one from other classes; or self-supervised setting~\cite{DBLP:conf/cvpr/WuXYL18,DBLP:conf/icml/ChenK0H20,DBLP:conf/cvpr/He0WXG20}, which select positive samples with data augmentation or co-occurrence. Different from both setting, we choose pos./neg. samples according to the groundtruth in a {\em supervised} way, yet we have {\em no pre-defined categories}. Furthermore, most aforementioned methods use single-modal samples, e.g., images, while we utilize cross-modal moment-sentence pairs. There are some video-language pre-training methods~\cite{DBLP:journals/corr/abs-2001-05691,DBLP:conf/cvpr/MiechASLSZ20} similar to our setting, yet they aim to learn {\em video-level} representations in an {\em unsupervised} way while we want to enhance {\em proposal-level} features in a {\em supervised} way, where no pre-training dataset is used. We adopt cross-modal pair discrimination loss~\cite{DBLP:journals/corr/abs-2001-05691} for our mutual matching scheme, where each instance in both modalities defines an unique class and the binary classes for cross-modal pairs, i.e., matched or unmatched, provide us valuable supervisions. Cross-modal image/video retrieval methods~\cite{DBLP:conf/cvpr/WangLL16,DBLP:conf/eccv/Gabeur0AS20} also utilize negative samples in a metric learning framework, yet they still treat the {\em image/video as a whole}. Thus we address a different problem from them.
\figOVERVIEW

\section{Model}
We propose to use negative sentence samples to construct a cross-modal mutual matching scheme for modeling bi-directional matching relations in visual and language modalities. To enable this auxiliary task, we adopt a late modality fusion from the metric-learning prospective with the advantage of less computational cost than the baseline~\cite{DBLP:conf/aaai/ZhangPFL20}. Sec.~\ref{sec:task} introduces the formulation of temporal grounding. Sec.~\ref{sec:arch} illustrates how to construct our MMN from the baseline. Then, we analyze the two complementary loss of IoU regression and mutual matching in Sec.~\ref{sec:loss}.

\subsection{Problem Formulation}
\label{sec:task}
Given an untrimmed video $V$ and a natural language query $S$, the temporal grounding task aims to localize a temporal moment $(x_s, x_e)$ that matches the query. We denote the video as a sequence of frames $V=\left\{x_{i}\right\}_{i=1}^{l_{v}}$, where $x_i$ is a frame and $l_v$ is the total number of frames; the query sentence as a sequence of words $S=\left\{s_{i}\right\}_{i=1}^{l_{s}}$, where $s_i$ is a word and $l_s$ is the total number of words. Ideally, the retrieved moment $(x_s^*, x_e^*)$ should deliver the same semantic as the sentence $S$. We adopt the feature vectors in the joint space to represent sentence $S$ and moments $(x_i, x_j)$, therefore the inner product of visual and text features after $\ell_2$-norm should be maximized.

\subsection{Architecture}
\label{sec:arch}
Our MMN adopts a Siamese-alike network architecture with a late modality fusion by a simple inner product in the joint visual-language space, as illustrated in Fig.~\ref{fig:overview}.

\noindent{\bf Language Encoder.} Previous works commonly utilize a LSTM~\cite{DBLP:journals/neco/HochreiterS97} upon a sequence of word vectors embedded by GloVe~\cite{DBLP:conf/emnlp/PenningtonSM14}. Yet, some concerns about unfair comparisons in previous methods, e.g., GloVe models pre-trained on different corpora, motivate us to adopt a standard language encoder. We choose DistilBERT~\cite{DBLP:journals/corr/abs-1910-01108} for its light-weighted model capacity. For each input sentence $S$, we first generate the tokens of words by the tokenizer and add a class embedding token `[CLS]' in the beginning. Then we feed the tokens into DistilBERT to get a feature sequence $\left\{\mathbf{f}^S_{i}\right\}_{i=1}^{l_s + 1}$, where $\mathbf{f}^S_{i} \in \mathbb{R}^{d^S}$ and $d^S$=768 is the feature dimension. There are two commonly used aggregation approaches to get the whole sentence's embedding: 1) global average pooling over all tokens
; 2) the class embedding `[CLS]'. Our experiment shows that global average pooling has faster convergence and better performance in this task, so we take average pooling as default. 



\noindent{\bf Video Encoder.} We extract features of the input video and encodes them as a 2D temporal moment feature map following the baseline 2D-TAN~\cite{DBLP:conf/aaai/ZhangPFL20}. We segment the input video stream into small video clips $\{v_i\}_{i=1}^{l_v / t}$ with each clip $v_i$ containing $t$ frames, then extract clip-level features with an off-the-shelf pre-trained CNN model (e.g., C3D). We perform a fixed-length sampling to obtain $N$ clip-level features for each video by an even stride $\frac{l_v}{t \cdot N}$ and pass the fixed-length features through an FC layer to reduce their dimension, denoted as $\{\mathbf{f}^V_i\}_{i=1}^{N}$, where $\mathbf{f}^V_i \in \mathbb{R}^{d^V}$. Then we build up the 2D feature map for candidate moments following the baseline~\cite{DBLP:conf/aaai/ZhangPFL20} as $\mathbf{F}^{M} \in \mathbb{R}^{N \times N \times d^{V}}$, where we adopt the max-pooling as the moment-level feature aggregation strategy following. We also utilize the sparse sampling strategy which removes highly overlapped moments to reduce the number of candidate moments as well as computation cost following 2D-TAN. Different from 2D-TAN~\cite{DBLP:conf/aaai/ZhangPFL20}, we directly model the relations of moments $\mathbf{F} \in \mathbb{R}^{N \times N \times d^{V}}$ by $L$ layers of 2D convolution with kernel size $K$ only based on visual features. As discussed in~\cite{DBLP:conf/bmvc/OtaniNRH20}, the performance of 2D-TAN has no significant drop when the order of input video clips is randomly permuted (i.e., the `sanity check'), which is counter-intuitive. It indicates that to some extent 2D-TAN ignores the visual features and overfits the bias in the dataset, such as temporal distributions of actions conditioned on sentences. The design of our MMN forces the convolution filters to actually utilize the visual features. 

\noindent{\bf Joint Visual-Language Embeddings.} Finally, we estimate the matching quality of each moment based on the similarity of two modalities for both supervision signals (i.e., IoU regression and cross-modal mutual matching). We adopt a LayerNorm~\cite{DBLP:journals/corr/BaKH16} for more stable convergence in language feature aggregation. We use a linear projection layer or a $1$x$1$ convolution to project the language and visual features into the same dimension $d^H$ respectively. The final representations of sentence feature are $\mathbf{f}_{mm}^S, \mathbf{f}_{iou}^S \in \mathbb{R}^{d^H}$ for cross-modal mutual matching (subscript $mm$) and IoU regression (subscript $iou$). Moment features are $\mathbf{F}_{mm}^V, \mathbf{F}_{iou}^V \in \mathbb{R}^{N \times N \times d^H}$. 
\begin{equation}
\mathbf{f}^S = \text{LayerNorm}(\mathbf{f}^S_1) \ \ \text{or} \ \ \text{LayerNorm}(\frac{1}{l_s}\sum_{i=2}^{l_s+1} \mathbf{f}^S_i)
\label{equ:sent-fc}
\end{equation}
\begin{equation}
\mathbf{f}_{mm}^S = \mathbf{W}_{mm}\mathbf{f}^S + \mathbf{b}_{mm}, \ \
\mathbf{f}_{iou}^S = \mathbf{W}_{iou}\mathbf{f}^S + \mathbf{b}_{iou} \\
\end{equation}
\begin{equation}
 \mathbf{F}_{mm}^V = \text{conv}_{mm}(\mathbf{F}, 1, 1), \ \ \mathbf{F}_{iou}^V =\text{conv}_{iou}(\mathbf{F}, 1, 1) \\
 \end{equation}
where $\mathbf{W}_{\cdot}$ and $\mathbf{b}_{\cdot}$ are learnable parameters, $\text{conv}_{\cdot}(\mathbf{x}, k, s)$ is 2D convolution with kernel size $k$ and stride $s$ for 2D feature map $\mathbf{x}$. The subscripts $id$ or $iou$ means the weights of two branches are independent. Then we regard the cosine similarity as moments' estimation scores for both losses. 

\begin{equation}
\begin{aligned}
\forall \mathbf{f}^V_{mm} \in \mathbf{F}^V_{mm}, s^{mm} = \mathbf{f}^{V \mathrm{T}}_{mm} \mathbf{f}_{mm}^S\\ 
\forall \mathbf{f}^V_{iou} \in \mathbf{F}^V_{iou}, s^{iou} = \mathbf{f}^{V \mathrm{T}}_{iou} \mathbf{f}_{iou}^S
\end{aligned}
\end{equation}
where we enforce the embedding $||\mathbf{f}_{\cdot}^V||_2 = ||\mathbf{f}_{\cdot}^S||_2 =1$ via a $\ell_2$-normalization layer. 

\subsection{Loss Functions}
\label{sec:loss}
Our MMN integrates two complementary losses: a binary cross entropy loss for regressing the IoU and a pair discrimination loss for learning discriminative features. 

\noindent{\bf Binary Cross Entropy.} We follow 2D-TAN~\cite{DBLP:conf/aaai/ZhangPFL20} to adopt the scaled IoU values $y_i$ for each candidate moments as the supervision signal, as shown in Tab.~\ref{tab:supervision} (IoU regression). The IoU values are linearly scaled from ($t_{min}$,$t_{max}$) to (0,1) and truncate the values beyond (0,1). We directly use the value of $t_{min}$and $t_{max}$ reported in 2D-TAN for fair comparisons. We notice that the range of cosine similarity is $s^{iou} \in$ (-1,1), yet the IoU signal is $y_i \in$ (0,1). So we adopt a commonly-used sigmoid function $\sigma$ to highlight the value change near the neutral region (e.g., $y_i$=0.5). We heuristically amplify the $s^{iou}$ by a factor of 10 to make the range of final prediction of our model $p^{iou}_i= \sigma(10 \cdot s_i^{iou})$ cover most of the regions in $(0,1)$. The regression branch of our MMN is trained by a BCE loss:
\begin{small}
\begin{equation}
L_{bce} =-\frac{1}{C} \sum_{i=1}^{C} \left( y_{i} \log p^{iou}_{i}+(1-y_{i}) \log (1-p^{iou}_{i}) \right),
\end{equation}
\end{small}
where $p^{iou}_i$ is the final score of a moment and $C$ is the total number of valid candidates.
\tableSUPERVISION

\noindent{\bf Cross-modal Mutual Matching.} As discussed in Sec.~\ref{sec:intro}, our cross-modal mutual matching creates more supervision signals for temporal grounding. By contrasting the positive moment-sentence pairs with the negative ones sampled from {\em both intra and inter videos}, encoders will learn a more discriminative features for both modalities {\em without any extra pre-training dataset}. As shown in Tab.~\ref{tab:supervision}, our mutual matching objective introduces two novel aspects of supervision signals: 1) pairs of the ground-truth moment and pos./neg. sentences (row 2); and 2) inter-video negative pairs (column 7). In contrast, previous detection/regression methods only sample pos./neg. samples based on IoU signals (row 3), and the previous metric learning method~\cite{DBLP:conf/iccv/HendricksWSSDR17} only has single direction of matching (i.e., only row 1) yet lacks the important cross-modal mutual matching (i.e., both row 1 and row 2). Specifically, we adapt the cross-modal pair discrimination loss~\cite{DBLP:journals/corr/abs-2001-05691} from video-level to proposal-level to learn features for moments $\mathbf{f}_{mm}^V$ and sentences $\mathbf{f}_{mm}^S$, where these features should be similar if the moment-sentence pair is semantically matched and dissimilar if it is semantically unrelated. We adopts the following conditional distribution in a non-parametric softmax form:
\begin{small}
\begin{equation}
\begin{aligned}
&p(i_s|v) = \frac{\text{exp}\left((\mathbf{f}_i^{S \mathrm{T}} \mathbf{f}^V -m)/\tau_v\right)}{\text{exp}\left((\mathbf{f}_i^{S \mathrm{T}} \mathbf{f}^V -m)/\tau_v\right) + \sum_{j\neq i}^{N_s} \text{exp} \left(\mathbf{f}_j^{S \mathrm{T}} \mathbf{f}^V /\tau_v\right)} \\
&p(i_v|s) = \frac{\text{exp}\left((\mathbf{f}_i^{V \mathrm{T}} \mathbf{f}^S -m)/\tau_s\right)}{\text{exp}\left((\mathbf{f}_i^{V \mathrm{T}} \mathbf{f}^S -m)/\tau_s\right) + \sum_{j\neq i}^{N_v} \text{exp} \left(\mathbf{f}_j^{V \mathrm{T}} \mathbf{f}^S /\tau_s\right)} \\
\end{aligned}
\end{equation}
\end{small}
where the $i^{th}$ sentence or moment define a instance-level class $i_s$ or $i_v$, the feature embedding $\mathbf{f}^S$ and $\mathbf{f}^V$ are $\ell_2$ normalized, $\tau_s$ and $\tau_v$ are temperatures and $N_s$ and $N_v$ are total numbers of sampled instances in the batch. Although the conditional distribution of video moments and sentences are similar, their differences are non-trivial due to the selection of negative samples, i.e., $N_v-1$ negative video moments are sampled from low IoU moments inside videos or moments from other videos; and $N_s-1$ negative sentences are sampled from other sentences in the video or from other videos. To enable a steady training process, we only adopt the moments whose IoU with ground-truth moment lower than a threshold as negative samples (e.g., $\leq$ 0.5). To further reduce the potential false negative signals, the sentences similar to the ground-truth sentence is automatically removed from the negative sample set, i.e., by computing their matched moments' IoU with ground-truth moment and removing them if their IoU $\geq$ 0.5. When we construct the negative sample set, we hold an assumption that only a small portion of inter-video negative samples (both sentences and moments) will semantically close to the positive sample. It is reasonable if the size of the training video corpus is large enough.

The objective of our cross-modal mutual matching is to maximize the likelihood $\prod^{N}_{i=1}p(i_s|v_i)\prod^{N}_{i=1}p(i_v|s_i)$ where $N$ is the total number of moment-sentence pairs for training. The cross-correlation enables the network to effectively capture the mutual information between modalities by guiding the learning process of feature representation learning with the binary pair supervision. The loss function is as follows
\begin{equation}
L_{mm} = - \left( \sum_{i=1}^{N} \log p(i_v|s_i) + \sum_{i=1}^{N} \log p(i_s|v_i)\right)
\end{equation}

Our final loss function $L$ is a linear combination of binary cross entropy loss and mutual matching loss, and the final prediction score $s$ for candidate moments given the query sentence are these two scores' product.
\begin{equation}
L = L_{bce} + \lambda L_{mm}, \quad s = s^{iou} \cdot s^{mm}.
\end{equation}

\section{Experiments}
\subsection{Datasets}
\noindent{\bf ActivityNet-Captions}~\cite{DBLP:conf/iccv/KrishnaHRFN17} is built on ActivityNet v1.3 dataset~\cite{DBLP:conf/cvpr/HeilbronEGN15}, where videos cover a wide range of complex human actions. It is originally designed for video captioning, and recently introduced into temporal grounding. There are 37,417, 17,505, and 17,031 moment-sentence pairs for training, validation, and testing respectively. Following the setting of 2D-TAN~\cite{DBLP:conf/aaai/ZhangPFL20}, we report the evaluation result on val\_2 set. 

\noindent{\bf TACoS} consists of 127 videos selected from the MPII-Cooking dataset~\cite{DBLP:conf/eccv/RohrbachRAAPS12}. It is comprised of 18,818 video-language pairs of different cooking activities in the kitchen annotated by \cite{DBLP:journals/tacl/RegneriRWTSP13}. A standard split~\cite{DBLP:conf/iccv/GaoSYN17} consists of 10,146, 4,589, and 4,083 moment-sentence pairs for training, validation and testing, respectively. We report evaluation results on test set in our experiments.

\noindent{\bf Charades-STA}~\cite{DBLP:conf/iccv/GaoSYN17} is an extended version of action recognition and localization dataset Charades~\cite{DBLP:conf/eccv/SigurdssonVWFLG16} by \cite{DBLP:conf/iccv/GaoSYN17} for temporal grounding. It contains 5,338 videos and 12,408 query-moment pairs in the training set, and 1,334 videos and 3,720 query-moment pairs in the test set.

\noindent{\bf HC-STVG} dataset is introduced by \cite{DBLP:journals/corr/abs-2011-05049} to spatio-temporally localize an action tubelet of the target person from an untrimmed video based on a given textual description. The dataset contains 5,660 videos selected from AVA~\cite{DBLP:conf/cvpr/GuSRVPLVTRSSM18} where each video has one video-sentence pair. A standard split consists of 4,500 pairs for training and 1,160 pairs for testing.

\subsection{Experimental Settings}
\label{sec:details}
\noindent{\bf Evaluation Metrics.} Following previous setting~\cite{DBLP:conf/iccv/GaoSYN17} of temporal grounding, we evaluate our model by computing {\it Rank n@m}. It is defined as the percentage of sentence queries having at least one correctly localized moment, i.e. IoU $\ge m$, in the top-$n$ retrieved moments. There are specific settings of $n$ and $m$ for different datasets. Specifically, we report the results as $m \in \{0.5, 0.7\}$ for Charades-STA, $m \in \{0.1, 0.3, 0.5\}$ for TACoS and $m \in \{0.3, 0.5, 0.7\}$ for ActivityNet Captions dataset with $n \in \{1, 5\}$. 

For spatio-temporal video grounding, we report the mean of vIoU and vIoU@\{0.3, 0.5\} following HC-STVG benchmark~\cite{DBLP:journals/corr/abs-2011-05049}, where vIoU is computed by the top-1 predicted tube with the GT tube. 

\noindent{\bf Implementation Details.} 
We adopt standard off-the-shelf video feature extractors without fine-tuning on each dataset. Our convolution network for 2D proposal feature modeling uses exactly the same settings with 2D-TAN~\cite{DBLP:conf/aaai/ZhangPFL20} (max-pool version) for fair comparisons, including visual features (VGG feature for Charades and C3D feature for ActivityNet-Captions and TACoS), number of sampled clips $N$, number of 2D convolution network layers $L$, kernel size $K$ and channels $d^{V}$, non maximum suppression (NMS) threshold, scaling thresholds $t_{min}$ and $t_{max}$. We set the dimension of the joint feature space $d^{H}=256$, and temperatures $\tau_s=\tau_v=0.1$. We use the HuggingFace~\cite{DBLP:journals/corr/abs-1910-03771} implementation of DistilBERT~\cite{DBLP:journals/corr/abs-1910-01108} with pre-trained model `distilbert-base-uncased' for better standardization in temporal grounding. In pair discrimination loss, we only sample negative moments lower than IoU=0.5 and set margin $m$ as $0.1$, $0.3$ and $0.4$ for TACoS, ActivityNet-Captions and Charades-STA respectively. We use AdamW~\cite{DBLP:conf/iclr/LoshchilovH19} optimizer with learning rate of $1 \times 10^{-4}$ and batch size $48$ for Charades, learning rate of $8 \times 10^{-4}$ and batch size $48$ for ActivityNet Captions and learning rate of $1.5 \times 10^{-3}$ and batch size $4$ for TACoS. We set $\lambda$ of pair discrimination loss as $0.1$ for ActivityNet Captions and $0.05$ for Charades and TACoS following the principle that both loss should contribute equally weighted gradients. We early stop the pair discrimination loss when we observe the performance on validation set starts to drop. The learning rate of DistilBERT is always $1/10$ of our main model. Each mini-batch has $B$ {\it videos} instead of $B$ {\it moment-sentence pairs}, where $B$ is batch size.
\tablePairDiscrimination
\tableNUMBER
\subsection{Ablation Study}
\label{sec:ablation}
\noindent{\bf Cross-modal Mutual Matching.} We investigate the most important part in our method, i.e., `cross-modal mutual matching', by ablating different types of negative samples on both Charades-STA and ActivityNet Captions datasets, as shown in Tab.~\ref{tab:pd}. Row 1 is the baseline of our model for not using mutual matching. Row 2 uses the same information with previous works (e.g., intra-video moment negatives), which shows similar performance with row 1. By introducing sentence negatives, the notable performance gain is observed in row 3 to row 6 when compared to row 1 and row 2. It proves the effectiveness of our mutual matching scheme. Besides, we conclude some observations based on the differences inside row 3 to row 8: (1) Due to the larger number of negatives, only using inter-video negatives (row 4) tend to be more effective than only using intra-video negatives (row 3), especially on smaller datasets (e.g., Charades-STA); (2) Best results are often obtained by both intra and inter-video negatives (row 8), and sub-optimal results are often achieved in row 5 and 6 by discarding some specific type of inter-video negatives. It indicates hard negative mining of inter-video samples could be further investigated; (3) Only using our mutual matching scheme achieves similar or better performance than original BCE loss on two datasets (row 7 vs. row 1), demonstrating the powerful additional supervision signals introduced by our mutual matching scheme.
\tableDistilBERT
\tableAGGREGATION
\figSANITY
\begin{figure}[t]
	\centering
	\includegraphics[height=4cm]{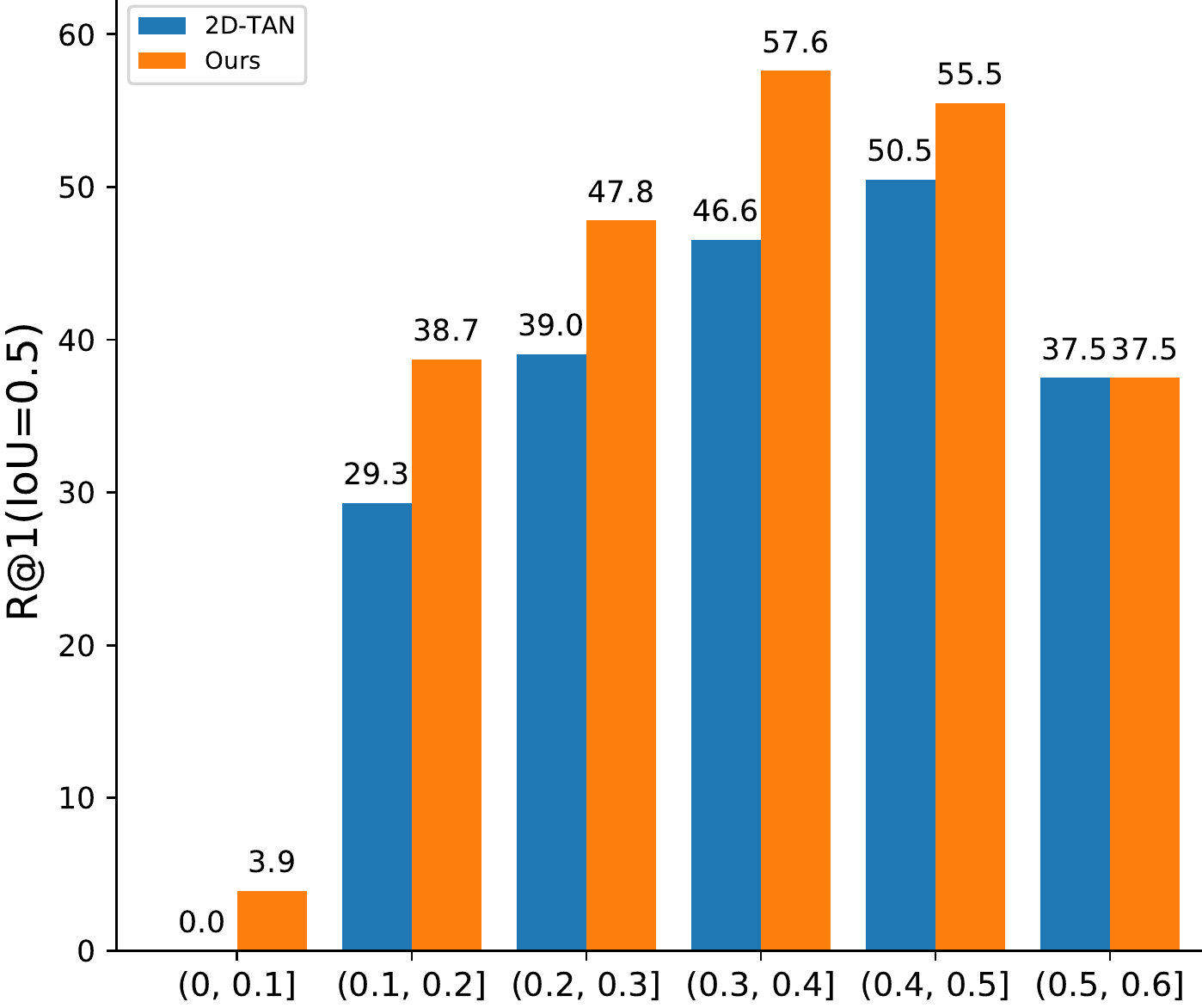}
	\caption{Performance comparisons decomposed by moment length on Charades-STA, R@1(IoU=0.5).}
	\label{fig:charades1}
\end{figure}

\noindent{\bf Number vs. Type of Negatives.} We have shown inter-video negatives are beneficial by introducing much more negatives than previous methods in Tab.~\ref{tab:pd}. However, whether the number or the type of negatives is more important in our method? To ablate this, we construct a baseline that randomly picks the same numbers of negatives with intra-video negatives among all intra and inter-video negatives (row 2). Therefore, it uses the {\bf same} number with only intra-video negatives (row 1) {\em in each iteration} yet will eventually see more types of negatives in the training process. Tab.~\ref{tab:number} shows that the performance of this baseline (row 2) is similar to only using intra-video negatives (row 1) in R@1 metrics yet is better in R@5 metrics, indicating the importance of more types of negatives introduced by our mutual matching scheme.

\noindent{\bf Previous Methods with DistilBERT}. Two baselines (DRN~\cite{DBLP:conf/cvpr/ZengXHCTG20} and 2D-TAN~\cite{DBLP:conf/aaai/ZhangPFL20}) with DistilBERT are constructed in Tab.~\ref{tab:bert}, which shows that the usage of DistilBERT has very little impact to the final performance of both methods. Therefore, we believe comparisons of MMN with previous SOTA methods are fair.

\noindent{\bf Aggregation of Text Feature.} In Tab.~\ref{tab:aggregation} (top), we compare two aggregation strategies for the sentence feature. While many methods in NLP including BERT~\cite{DBLP:conf/naacl/DevlinCLT19} directly adopt the class embedding as the representation of the whole sentence, we find the average pooling over all words is more robust in our task. It indicates that the scale of video grounding datasets is limited in vocabulary size or annotated sentences, e.g., TACoS dataset only has a vocabulary size of 2255 words and a training set of 10146 sentences, while NLP pre-training datasets commonly have millions of sentences. We empirically use average pooling as default.

\noindent{\bf Hyper-parameters.} In Tab.~\ref{tab:aggregation} (bottom), we ablate important hyper-parameters of pair discrimination loss like margin $m$, temperature $\tau_s,\tau_v$ and weight $\lambda$. Due to the small size of Charades-STA, they do have some influences but no notable change of the gap between our MMN and baseline~\cite{DBLP:conf/aaai/ZhangPFL20}. We set them by just a few tries on each dataset.

\label{sanity-check}
\noindent{\bf Sanity Check on Visual Input.} \cite{DBLP:conf/bmvc/OtaniNRH20} observe that the blind baselines which only uses training set priors or conditional distributions of moments given queries already outperform many previous SOTA methods. So they define a test named `sanity check' to examine the contribution of visual inputs by evaluating the performance gap between the ordered input clips and randomly shuffled ones. We follow this setting on Charades-STA in Fig.~\ref{fig:sanity}: Our model with randomized visual input predicts an even lower result (37.5 vs. 37.8) than Blind-TAN~\cite{DBLP:conf/bmvc/OtaniNRH20} which does not use any visual input. Our performance gap between ordered inputs and shuffled ones is also larger than 2D-TAN's, which means our MMN shows better result in this sanity check. It indicates our method better exploits visual information. 
\tableCHARADES
\tableANET
\tableTACOS

\noindent{\bf Performance Comparisons by Moment Length.} To further discover our improvement over the baseline~\cite{DBLP:conf/aaai/ZhangPFL20}, we decompose the performance comparisons by the ratio of moment length over the video length in Fig.~\ref{fig:charades1}. The x-axis is grouped by the moment length ratio which stops at 0.6 because the maximum length of moments on Charades-STA is less than 0.6. It is worth noticing that the number of moments in each ratio interval is not balanced, e.g., GT moments have 1020 instances in (0.1, 0.2] while only 16 in (0.5, 0.6]. Our MMN consistently outperforms 2D-TAN in various moment lengths.
\figVISUAL
\subsection{Comparison with the State of the Art} 
We achieve highly competitive results on three datasets in temporal grounding: ActivityNet Captions, Charades-STA and TACoS, as reported in Tab.~\ref{tab:anet}, Tab.~\ref{tab:charades}, and Tab.~\ref{tab:tacos}. The top-2 performance values in tables are highlighted by {\bf bold} or \underline{ underline}. Based on the results, we have several observations. {\bf Firstly}, our proposed MMN outperforms the strong baseline~\cite{DBLP:conf/aaai/ZhangPFL20} with a significant margin by leveraging our mutual matching scheme (i.e., using both video and sentence negative samples from both intra and inter videos). As a result, it outperforms or is on par with most of the previous or recently proposed SOTA methods. Specifically, we compare our MMN with several concurrent approaches which share the similar loss function or architecture with ours, i.e., IVG-DCL~\cite{DBLP:conf/cvpr/NanQXLLZL21},  SSCS~\cite{Ding_2021_ICCV} and FVMR~\cite{Gao_2021_ICCV}. IVG-DCL and SSCS use contrastive loss to align the sentences and video {\em clips} instead of the sentences and {\em moments} in MMN. Thus they lack our important auxiliary task `mutual matching' and ignore the important sentence negative samples. They are also more complex than ours due to the additional IVG module (in IVG-DCL) or captioning/support-set module (in SSCS). As a result, they achieve similar or worse results than ours. They also do not address temporal grounding from a metric-learning perspective, thus their proposed modules will only add more computations to their baselines instead of reducing them, which is different from ours. FVMR also address temporal grounding from a metric-learning perspective, yet our supervision signals are totally different: 1) FVMR adopts a knowledge distillation loss where the teacher is the original hadamard product of cross-modal features for fusion with a MLP for prediction and the student is the inner-product in the common space. Our MMN adopts a contrastive loss and directly uses a inner-product in the common space without other fusion strategies as guidance. 2) FVMR also ignores the important sentence negative samples and lacks our proposed `mutual matching scheme', thus it uses the same supervisions as 2D-TAN and fails to construct more supervisions like ours. Besides, FVMR utilizes a complex semantic role tree for extracting phrase-level fine-grained sentence feature, while we use a simple average pooling over all words to obtain our sentence features. We achieve a better performance than FVMR on two datasets out of three, including the largest ActivityNet. 
{\bf Secondly}, it is worth noting that our framework is independent to the moment-relation modeling network or language encoders, so our MMN could be implemented in any proposal-based temporal grounding methods and further boost the performances on strong baselines, e.g., MS-2D-TAN~\cite{DBLP:journals/corr/abs-2012-02646} with multi-scale modelling capacity. 
{\bf Thirdly}, our computational cost is significantly reduced compared with 2D-TAN~\cite{DBLP:conf/aaai/ZhangPFL20} due to shared video features between sentences inside each video. The training process of the baseline takes 36 GPU hours to converge on ActivityNet-Captions dataset while ours only takes 10 GPU hours on the same GPU server. The testing cost could also be reduced if there are multiple sentences in a video, e.g., there are hundreds of sentences per video in TACoS dataset.

\noindent{\bf Visualizations.} In Fig.~\ref{fig:visual_anet_charades}, we show the visualizations on test set of two datasets. Invalid regions of the first three maps are manually set to be in color `grey'. Due to highly unbalanced pos./neg. samples of BCE loss in 2D-TAN~\cite{DBLP:conf/aaai/ZhangPFL20}, the absolute value of top-1 result is lower than our expectation. Yet, the gap between top-1 prediction and backgrounds is still discriminative for localizing moments. We conclude that IoU scores tend to be sharp near the GT location yet could be wrong in many situations, while pair discrimination scores tend to cover the right location yet are not sharp among hard negatives. These two score maps are complementary to each other. In our MMN, IoU regression branch is already improved by sharing the same backbone enhanced by pair discrimination loss. The final performance could be further improved a bit (about $0.5\%$) by fusing their scores with a simple product in our experiments. In the last map, we use t-SNE~\cite{JMLR:v9:vandermaaten08a} to visualize features in the joint space ($d^H$=256) to a 2D image. Based on the distribution of moments, we believe our joint space has learnt an effective similarity measurement for both video and language modalities.

\subsection{Spatio-Temporal Video Grounding}
We also extend MMN to spatio-temporally localize action tubes given queries to further show its generalization ability. We find that MMN is even more effective on the HC-STVG dataset due to its smaller scale, where our additional supervisions are very important for limited annotations.

\noindent{\bf Adapting MMN to the STVG task.} We adopt a two-stage method for this task: In stage 1, we first detect human bounding boxes by Faster R-CNN~\cite{DBLP:conf/nips/RenHGS15} and link them to be candidate tubes following ACT~\cite{DBLP:conf/iccv/KalogeitonWFS17a}; then we use CSN~\cite{DBLP:conf/iccv/TranWFT19} to extract 1024-d RoI features for each bounding box to form a feature sequence per candidate tube. In stage 2, we use our MMN to temporally refine the candidate tubes with proper adaptation. There are several candidate tubes (as feature sequences) generated in stage 1 in each video, while temporal grounding only have one feature sequence per video. Thus, we generate a 2D moment map for each candidate tube, yet there is only one positive moment per video selected by the highest vIoU globally and all the rest moments including other 2D moment maps are negatives. The IoU regression branch also uses the scaled vIoU as the groundtruth. As a result, vIoU-based supervisions will automatically guide our MMN to find the right tube and the appropriate temporal refinement and we do not need to design complex evaluation module (e.g., transformer~\cite{DBLP:journals/corr/abs-2011-05049}) to select the right candidate tube to be refined.

\noindent{\bf Performance Comparisons.} In Tab.~\ref{table:stvg}, our MMN shows significant performance gain over some baselines or recently proposed transformer-based methods, e.g., 2D-TAN+WSSTG~\cite{DBLP:conf/acl/ChenMLW19}, STGVT~\cite{DBLP:journals/corr/abs-2011-05049} and STVGBert~\cite{Su_2021_ICCV}. Our MMN also shows notable performance gain over 2D-TAN~\cite{DBLP:conf/aaai/ZhangPFL20} on the same candidate tubes generated from our stage 1. Finally, our method outperforms some competitive opponents~\cite{DBLP:journals/corr/abs-2106-07166,DBLP:journals/corr/abs-2106-10634} which use strong multi-modal pre-training models (e.g., LXMERT~\cite{DBLP:conf/emnlp/TanB19} or MDETR~\cite{DBLP:journals/corr/abs-2104-12763}) and ranks first in the HC-STVG challenge of the 3rd PIC workshop\footnote{\url{http://www.picdataset.com/challenge/task/hcvg/}}. In contrast, cross-modal interaction modeling in our method is solely achieved by our MMN and no multi-modal pre-training model on extra large-scale video datasets is used.
\tableSTVG

\section{Conclusion}
In this paper, we propose the Mutual Matching Network (MMN) in the metric-learning prospective for temporal grounding. Particularly, we first propose to use the auxiliary task of {\em mutual matching} which asks the model to select the correct sentence in a constructed negative sentence set for video moments in addition to existing supervisions. By leveraging the powerful textual negatives, more discriminative features for both modalities are learned by cross-modal mutual matching. Moreover, inter-video negatives in temporal grounding are effectively exploited for the first time in our method. With this simple, effective and efficient framework, we achieve state-of-the-art performance on four challenging video grounding benchmarks: Charades-STA, TACoS, ActivityNet-Captions, and HC-STVG.

\section*{Acknowledgements} 
This work is supported by the National Science Foundation of China (No. 62076119, No. 61921006), Program for Innovative Talents and Entrepreneur in Jiangsu Province, and Collaborative Innovation Center of Novel Software Technology and Industrialization. Authors would like to thank Yixuan Li for her effort in the HC-STVG challenge.
\clearpage
\bibliographystyle{aaai22}
\bibliography{aaai22}

\begin{thebibliography}{74}
\providecommand{\natexlab}[1]{#1}

\bibitem[{Ba, Kiros, and Hinton(2016)}]{DBLP:journals/corr/BaKH16}
Ba, L.~J.; Kiros, J.~R.; and Hinton, G.~E. 2016.
\newblock Layer Normalization.
\newblock \emph{CoRR}, abs/1607.06450.

\bibitem[{Chen et~al.(2018)Chen, Chen, Ma, Jie, and
  Chua}]{DBLP:conf/emnlp/ChenCMJC18}
Chen, J.; Chen, X.; Ma, L.; Jie, Z.; and Chua, T. 2018.
\newblock Temporally Grounding Natural Sentence in Video.
\newblock In \emph{{EMNLP}}.

\bibitem[{Chen et~al.(2020{\natexlab{a}})Chen, Lu, Tang, Xiao, Zhang, Tan, and
  Li}]{DBLP:conf/aaai/ChenLTXZTL20}
Chen, L.; Lu, C.; Tang, S.; Xiao, J.; Zhang, D.; Tan, C.; and Li, X.
  2020{\natexlab{a}}.
\newblock Rethinking the Bottom-Up Framework for Query-Based Video
  Localization.
\newblock In \emph{{AAAI}}.

\bibitem[{Chen et~al.(2020{\natexlab{b}})Chen, Jiang, Liu, and
  Jiang}]{DBLP:conf/eccv/ChenJLJ20}
Chen, S.; Jiang, W.; Liu, W.; and Jiang, Y. 2020{\natexlab{b}}.
\newblock Learning Modality Interaction for Temporal Sentence Localization and
  Event Captioning in Videos.
\newblock In \emph{{ECCV}}.

\bibitem[{Chen and Jiang(2019)}]{DBLP:conf/aaai/ChenJ19a}
Chen, S.; and Jiang, Y. 2019.
\newblock Semantic Proposal for Activity Localization in Videos via Sentence
  Query.
\newblock In \emph{{AAAI}}.

\bibitem[{Chen et~al.(2020{\natexlab{c}})Chen, Kornblith, Norouzi, and
  Hinton}]{DBLP:conf/icml/ChenK0H20}
Chen, T.; Kornblith, S.; Norouzi, M.; and Hinton, G.~E. 2020{\natexlab{c}}.
\newblock A Simple Framework for Contrastive Learning of Visual
  Representations.
\newblock In \emph{{ICML}}.

\bibitem[{Chen et~al.(2019)Chen, Ma, Luo, and Wong}]{DBLP:conf/acl/ChenMLW19}
Chen, Z.; Ma, L.; Luo, W.; and Wong, K.~K. 2019.
\newblock Weakly-Supervised Spatio-Temporally Grounding Natural Sentence in
  Video.
\newblock In \emph{{ACL} {(1)}}.

\bibitem[{Devlin et~al.(2019)Devlin, Chang, Lee, and
  Toutanova}]{DBLP:conf/naacl/DevlinCLT19}
Devlin, J.; Chang, M.; Lee, K.; and Toutanova, K. 2019.
\newblock {BERT:} Pre-training of Deep Bidirectional Transformers for Language
  Understanding.
\newblock In \emph{{NAACL-HLT} {(1)}}.

\bibitem[{Ding et~al.(2021)Ding, Wang, Zhang, Cheng, Li, Huang, Tang, and
  Gao}]{Ding_2021_ICCV}
Ding, X.; Wang, N.; Zhang, S.; Cheng, D.; Li, X.; Huang, Z.; Tang, M.; and Gao,
  X. 2021.
\newblock Support-Set Based Cross-Supervision for Video Grounding.
\newblock In \emph{ICCV}.

\bibitem[{Feichtenhofer et~al.(2019)Feichtenhofer, Fan, Malik, and
  He}]{DBLP:conf/iccv/Feichtenhofer0M19}
Feichtenhofer, C.; Fan, H.; Malik, J.; and He, K. 2019.
\newblock SlowFast Networks for Video Recognition.
\newblock In \emph{{ICCV}}.

\bibitem[{Gabeur et~al.(2020)Gabeur, Sun, Alahari, and
  Schmid}]{DBLP:conf/eccv/Gabeur0AS20}
Gabeur, V.; Sun, C.; Alahari, K.; and Schmid, C. 2020.
\newblock Multi-modal Transformer for Video Retrieval.
\newblock In \emph{{ECCV} {(4)}}.

\bibitem[{Gao et~al.(2017{\natexlab{a}})Gao, Sun, Yang, and
  Nevatia}]{DBLP:conf/iccv/GaoSYN17}
Gao, J.; Sun, C.; Yang, Z.; and Nevatia, R. 2017{\natexlab{a}}.
\newblock {TALL:} Temporal Activity Localization via Language Query.
\newblock In \emph{{ICCV}}.

\bibitem[{Gao and Xu(2021)}]{Gao_2021_ICCV}
Gao, J.; and Xu, C. 2021.
\newblock Fast Video Moment Retrieval.
\newblock In \emph{ICCV}.

\bibitem[{Gao et~al.(2017{\natexlab{b}})Gao, Yang, Sun, Chen, and
  Nevatia}]{DBLP:conf/iccv/GaoYSCN17}
Gao, J.; Yang, Z.; Sun, C.; Chen, K.; and Nevatia, R. 2017{\natexlab{b}}.
\newblock {TURN} {TAP:} Temporal Unit Regression Network for Temporal Action
  Proposals.
\newblock In \emph{{ICCV}}.

\bibitem[{Ge et~al.(2019)Ge, Gao, Chen, and Nevatia}]{DBLP:conf/wacv/GeGCN19}
Ge, R.; Gao, J.; Chen, K.; and Nevatia, R. 2019.
\newblock {MAC:} Mining Activity Concepts for Language-Based Temporal
  Localization.
\newblock In \emph{{WACV}}.

\bibitem[{Ghosh et~al.(2019)Ghosh, Agarwal, Parekh, and
  Hauptmann}]{DBLP:conf/naacl/GhoshAPH19}
Ghosh, S.; Agarwal, A.; Parekh, Z.; and Hauptmann, A.~G. 2019.
\newblock ExCL: Extractive Clip Localization Using Natural Language
  Descriptions.
\newblock In \emph{{NAACL-HLT} {(1)}}.

\bibitem[{Gu et~al.(2018)Gu, Sun, Ross, Vondrick, Pantofaru, Li,
  Vijayanarasimhan, Toderici, Ricco, Sukthankar, Schmid, and
  Malik}]{DBLP:conf/cvpr/GuSRVPLVTRSSM18}
Gu, C.; Sun, C.; Ross, D.~A.; Vondrick, C.; Pantofaru, C.; Li, Y.;
  Vijayanarasimhan, S.; Toderici, G.; Ricco, S.; Sukthankar, R.; Schmid, C.;
  and Malik, J. 2018.
\newblock {AVA:} {A} Video Dataset of Spatio-Temporally Localized Atomic Visual
  Actions.
\newblock In \emph{{CVPR}}.

\bibitem[{Hadsell, Chopra, and LeCun(2006)}]{DBLP:conf/cvpr/HadsellCL06}
Hadsell, R.; Chopra, S.; and LeCun, Y. 2006.
\newblock Dimensionality Reduction by Learning an Invariant Mapping.
\newblock In \emph{{CVPR} {(2)}}.

\bibitem[{He et~al.(2019)He, Zhao, Huang, Li, Liu, and
  Wen}]{DBLP:conf/aaai/HeZHLLW19}
He, D.; Zhao, X.; Huang, J.; Li, F.; Liu, X.; and Wen, S. 2019.
\newblock Read, Watch, and Move: Reinforcement Learning for Temporally
  Grounding Natural Language Descriptions in Videos.
\newblock In \emph{{AAAI}}.

\bibitem[{He et~al.(2020)He, Fan, Wu, Xie, and
  Girshick}]{DBLP:conf/cvpr/He0WXG20}
He, K.; Fan, H.; Wu, Y.; Xie, S.; and Girshick, R.~B. 2020.
\newblock Momentum Contrast for Unsupervised Visual Representation Learning.
\newblock In \emph{{CVPR}}.

\bibitem[{Heilbron et~al.(2015)Heilbron, Escorcia, Ghanem, and
  Niebles}]{DBLP:conf/cvpr/HeilbronEGN15}
Heilbron, F.~C.; Escorcia, V.; Ghanem, B.; and Niebles, J.~C. 2015.
\newblock ActivityNet: {A} large-scale video benchmark for human activity
  understanding.
\newblock In \emph{{CVPR}}.

\bibitem[{Hendricks et~al.(2017)Hendricks, Wang, Shechtman, Sivic, Darrell, and
  Russell}]{DBLP:conf/iccv/HendricksWSSDR17}
Hendricks, L.~A.; Wang, O.; Shechtman, E.; Sivic, J.; Darrell, T.; and Russell,
  B.~C. 2017.
\newblock Localizing Moments in Video with Natural Language.
\newblock In \emph{{ICCV}}.

\bibitem[{Hochreiter and Schmidhuber(1997)}]{DBLP:journals/neco/HochreiterS97}
Hochreiter, S.; and Schmidhuber, J. 1997.
\newblock Long Short-Term Memory.
\newblock \emph{Neural Comput.}, 9(8).

\bibitem[{Kalogeiton et~al.(2017)Kalogeiton, Weinzaepfel, Ferrari, and
  Schmid}]{DBLP:conf/iccv/KalogeitonWFS17a}
Kalogeiton, V.; Weinzaepfel, P.; Ferrari, V.; and Schmid, C. 2017.
\newblock Action Tubelet Detector for Spatio-Temporal Action Localization.
\newblock In \emph{{ICCV}}.

\bibitem[{Kamath et~al.(2021)Kamath, Singh, LeCun, Misra, Synnaeve, and
  Carion}]{DBLP:journals/corr/abs-2104-12763}
Kamath, A.; Singh, M.; LeCun, Y.; Misra, I.; Synnaeve, G.; and Carion, N. 2021.
\newblock {MDETR} - Modulated Detection for End-to-End Multi-Modal
  Understanding.
\newblock \emph{CoRR}, abs/2104.12763.

\bibitem[{Khosla et~al.(2020)Khosla, Teterwak, Wang, Sarna, Tian, Isola,
  Maschinot, Liu, and Krishnan}]{DBLP:conf/nips/KhoslaTWSTIMLK20}
Khosla, P.; Teterwak, P.; Wang, C.; Sarna, A.; Tian, Y.; Isola, P.; Maschinot,
  A.; Liu, C.; and Krishnan, D. 2020.
\newblock Supervised Contrastive Learning.
\newblock In \emph{NeurIPS}.

\bibitem[{Krishna et~al.(2017)Krishna, Hata, Ren, Fei{-}Fei, and
  Niebles}]{DBLP:conf/iccv/KrishnaHRFN17}
Krishna, R.; Hata, K.; Ren, F.; Fei{-}Fei, L.; and Niebles, J.~C. 2017.
\newblock Dense-Captioning Events in Videos.
\newblock In \emph{{ICCV}}.

\bibitem[{Li and Wang(2020)}]{DBLP:journals/corr/abs-2001-05691}
Li, T.; and Wang, L. 2020.
\newblock Learning Spatiotemporal Features via Video and Text Pair
  Discrimination.
\newblock \emph{CoRR}, abs/2001.05691.

\bibitem[{Li et~al.(2020)Li, Wang, Wang, and Wu}]{DBLP:conf/eccv/LiW0W20}
Li, Y.; Wang, Z.; Wang, L.; and Wu, G. 2020.
\newblock Actions as Moving Points.
\newblock In \emph{{ECCV}}.

\bibitem[{Lin et~al.(2019)Lin, Liu, Li, Ding, and
  Wen}]{DBLP:conf/iccv/LinLLDW19}
Lin, T.; Liu, X.; Li, X.; Ding, E.; and Wen, S. 2019.
\newblock {BMN:} Boundary-Matching Network for Temporal Action Proposal
  Generation.
\newblock In \emph{{ICCV}}.

\bibitem[{Lin et~al.(2020)Lin, Zhao, Zhang, Zhang, and
  Cai}]{DBLP:journals/tip/LinZZZC20}
Lin, Z.; Zhao, Z.; Zhang, Z.; Zhang, Z.; and Cai, D. 2020.
\newblock Moment Retrieval via Cross-Modal Interaction Networks With Query
  Reconstruction.
\newblock \emph{{IEEE} Trans. Image Process.}, 29.

\bibitem[{Liu et~al.(2018)Liu, Yeung, Chou, Huang, Fei{-}Fei, and
  Niebles}]{DBLP:conf/eccv/LiuYCHFN18}
Liu, B.; Yeung, S.; Chou, E.; Huang, D.; Fei{-}Fei, L.; and Niebles, J.~C.
  2018.
\newblock Temporal Modular Networks for Retrieving Complex Compositional
  Activities in Videos.
\newblock In \emph{{ECCV}}.

\bibitem[{Liu et~al.(2021)Liu, Qu, Dong, Zhou, Cheng, Wei, Xu, and
  Xie}]{DBLP:conf/cvpr/LiuQDZ00XX21}
Liu, D.; Qu, X.; Dong, J.; Zhou, P.; Cheng, Y.; Wei, W.; Xu, Z.; and Xie, Y.
  2021.
\newblock Context-Aware Biaffine Localizing Network for Temporal Sentence
  Grounding.
\newblock In \emph{{CVPR}}.

\bibitem[{Loshchilov and Hutter(2019)}]{DBLP:conf/iclr/LoshchilovH19}
Loshchilov, I.; and Hutter, F. 2019.
\newblock Decoupled Weight Decay Regularization.
\newblock In \emph{{ICLR}}.

\bibitem[{Lu et~al.(2019)Lu, Chen, Tan, Li, and
  Xiao}]{DBLP:conf/emnlp/LuCTLX19}
Lu, C.; Chen, L.; Tan, C.; Li, X.; and Xiao, J. 2019.
\newblock {DEBUG:} {A} Dense Bottom-Up Grounding Approach for Natural Language
  Video Localization.
\newblock In \emph{{EMNLP/IJCNLP} {(1)}}.

\bibitem[{Miech et~al.(2020)Miech, Alayrac, Smaira, Laptev, Sivic, and
  Zisserman}]{DBLP:conf/cvpr/MiechASLSZ20}
Miech, A.; Alayrac, J.; Smaira, L.; Laptev, I.; Sivic, J.; and Zisserman, A.
  2020.
\newblock End-to-End Learning of Visual Representations From Uncurated
  Instructional Videos.
\newblock In \emph{{CVPR}}.

\bibitem[{Mun, Cho, and Han(2020)}]{DBLP:conf/cvpr/MunCH20}
Mun, J.; Cho, M.; and Han, B. 2020.
\newblock Local-Global Video-Text Interactions for Temporal Grounding.
\newblock In \emph{{CVPR}}.

\bibitem[{Nan et~al.(2021)Nan, Qiao, Xiao, Liu, Leng, Zhang, and
  Lu}]{DBLP:conf/cvpr/NanQXLLZL21}
Nan, G.; Qiao, R.; Xiao, Y.; Liu, J.; Leng, S.; Zhang, H.; and Lu, W. 2021.
\newblock Interventional Video Grounding With Dual Contrastive Learning.
\newblock In \emph{{CVPR}}.

\bibitem[{Opazo et~al.(2020)Opazo, Marrese{-}Taylor, Saleh, Li, and
  Gould}]{DBLP:conf/wacv/OpazoMSLG20}
Opazo, C.~R.; Marrese{-}Taylor, E.; Saleh, F.~S.; Li, H.; and Gould, S. 2020.
\newblock Proposal-free Temporal Moment Localization of a Natural-Language
  Query in Video using Guided Attention.
\newblock In \emph{{WACV}}.

\bibitem[{Otani et~al.(2020)Otani, Nakashima, Rahtu, and
  Heikkil{\"{a}}}]{DBLP:conf/bmvc/OtaniNRH20}
Otani, M.; Nakashima, Y.; Rahtu, E.; and Heikkil{\"{a}}, J. 2020.
\newblock Uncovering Hidden Challenges in Query-Based Video Moment Retrieval.
\newblock In \emph{{BMVC}}.

\bibitem[{Pennington, Socher, and
  Manning(2014)}]{DBLP:conf/emnlp/PenningtonSM14}
Pennington, J.; Socher, R.; and Manning, C.~D. 2014.
\newblock Glove: Global Vectors for Word Representation.
\newblock In \emph{{EMNLP}}.

\bibitem[{Ramanishka et~al.(2017)Ramanishka, Das, Zhang, and
  Saenko}]{DBLP:conf/cvpr/RamanishkaDZS17}
Ramanishka, V.; Das, A.; Zhang, J.; and Saenko, K. 2017.
\newblock Top-Down Visual Saliency Guided by Captions.
\newblock In \emph{{CVPR}}.

\bibitem[{Regneri et~al.(2013)Regneri, Rohrbach, Wetzel, Thater, Schiele, and
  Pinkal}]{DBLP:journals/tacl/RegneriRWTSP13}
Regneri, M.; Rohrbach, M.; Wetzel, D.; Thater, S.; Schiele, B.; and Pinkal, M.
  2013.
\newblock Grounding Action Descriptions in Videos.
\newblock \emph{Trans. Assoc. Comput. Linguistics}, 1.

\bibitem[{Ren et~al.(2015)Ren, He, Girshick, and Sun}]{DBLP:conf/nips/RenHGS15}
Ren, S.; He, K.; Girshick, R.~B.; and Sun, J. 2015.
\newblock Faster {R-CNN:} Towards Real-Time Object Detection with Region
  Proposal Networks.
\newblock In \emph{{NIPS}}.

\bibitem[{Rohrbach et~al.(2012)Rohrbach, Regneri, Andriluka, Amin, Pinkal, and
  Schiele}]{DBLP:conf/eccv/RohrbachRAAPS12}
Rohrbach, M.; Regneri, M.; Andriluka, M.; Amin, S.; Pinkal, M.; and Schiele, B.
  2012.
\newblock Script Data for Attribute-Based Recognition of Composite Activities.
\newblock In \emph{{ECCV}}.

\bibitem[{Sanh et~al.(2019)Sanh, Debut, Chaumond, and
  Wolf}]{DBLP:journals/corr/abs-1910-01108}
Sanh, V.; Debut, L.; Chaumond, J.; and Wolf, T. 2019.
\newblock DistilBERT, a distilled version of {BERT:} smaller, faster, cheaper
  and lighter.
\newblock \emph{CoRR}, abs/1910.01108.

\bibitem[{Sigurdsson et~al.(2016)Sigurdsson, Varol, Wang, Farhadi, Laptev, and
  Gupta}]{DBLP:conf/eccv/SigurdssonVWFLG16}
Sigurdsson, G.~A.; Varol, G.; Wang, X.; Farhadi, A.; Laptev, I.; and Gupta, A.
  2016.
\newblock Hollywood in Homes: Crowdsourcing Data Collection for Activity
  Understanding.
\newblock In \emph{{ECCV}}.

\bibitem[{Su, Yu, and Xu(2021)}]{Su_2021_ICCV}
Su, R.; Yu, Q.; and Xu, D. 2021.
\newblock STVGBert: A Visual-Linguistic Transformer Based Framework for
  Spatio-Temporal Video Grounding.
\newblock In \emph{ICCV}.

\bibitem[{Tan et~al.(2021)Tan, Lin, Hu, Li, and
  Zheng}]{DBLP:journals/corr/abs-2106-10634}
Tan, C.; Lin, Z.; Hu, J.; Li, X.; and Zheng, W. 2021.
\newblock Augmented 2D-TAN: {A} Two-stage Approach for Human-centric
  Spatio-Temporal Video Grounding.
\newblock \emph{CoRR}, abs/2106.10634.

\bibitem[{Tan and Bansal(2019)}]{DBLP:conf/emnlp/TanB19}
Tan, H.; and Bansal, M. 2019.
\newblock {LXMERT:} Learning Cross-Modality Encoder Representations from
  Transformers.
\newblock In \emph{{EMNLP/IJCNLP} {(1)}}.

\bibitem[{Tang et~al.(2020)Tang, Liao, Liu, Li, Jin, Jiang, Yu, and
  Xu}]{DBLP:journals/corr/abs-2011-05049}
Tang, Z.; Liao, Y.; Liu, S.; Li, G.; Jin, X.; Jiang, H.; Yu, Q.; and Xu, D.
  2020.
\newblock Human-centric Spatio-Temporal Video Grounding With Visual
  Transformers.
\newblock \emph{CoRR}, abs/2011.05049.

\bibitem[{Tran et~al.(2019)Tran, Wang, Feiszli, and
  Torresani}]{DBLP:conf/iccv/TranWFT19}
Tran, D.; Wang, H.; Feiszli, M.; and Torresani, L. 2019.
\newblock Video Classification With Channel-Separated Convolutional Networks.
\newblock In \emph{{ICCV}}.

\bibitem[{van~der Maaten and Hinton(2008)}]{JMLR:v9:vandermaaten08a}
van~der Maaten, L.; and Hinton, G. 2008.
\newblock Visualizing Data using t-SNE.
\newblock \emph{Journal of Machine Learning Research}, 9(86).

\bibitem[{Wang, Ma, and Jiang(2020)}]{DBLP:conf/aaai/Wang0J20}
Wang, J.; Ma, L.; and Jiang, W. 2020.
\newblock Temporally Grounding Language Queries in Videos by Contextual
  Boundary-Aware Prediction.
\newblock In \emph{{AAAI}}.

\bibitem[{Wang et~al.(2018)Wang, Li, Li, and Gool}]{DBLP:conf/cvpr/WangL0G18}
Wang, L.; Li, W.; Li, W.; and Gool, L.~V. 2018.
\newblock Appearance-and-Relation Networks for Video Classification.
\newblock In \emph{{CVPR}}.

\bibitem[{Wang, Li, and Lazebnik(2016)}]{DBLP:conf/cvpr/WangLL16}
Wang, L.; Li, Y.; and Lazebnik, S. 2016.
\newblock Learning Deep Structure-Preserving Image-Text Embeddings.
\newblock In \emph{{CVPR}}.

\bibitem[{Wang et~al.(2021)Wang, Tong, Ji, and Wu}]{DBLP:conf/cvpr/0002TJW21}
Wang, L.; Tong, Z.; Ji, B.; and Wu, G. 2021.
\newblock {TDN:} Temporal Difference Networks for Efficient Action Recognition.
\newblock In \emph{{CVPR}}.

\bibitem[{Wang et~al.(2016)Wang, Xiong, Wang, Qiao, Lin, Tang, and
  Gool}]{DBLP:conf/eccv/WangXW0LTG16}
Wang, L.; Xiong, Y.; Wang, Z.; Qiao, Y.; Lin, D.; Tang, X.; and Gool, L.~V.
  2016.
\newblock Temporal Segment Networks: Towards Good Practices for Deep Action
  Recognition.
\newblock In \emph{{ECCV}}.

\bibitem[{Wang, Huang, and Wang(2019)}]{DBLP:conf/cvpr/WangHW19}
Wang, W.; Huang, Y.; and Wang, L. 2019.
\newblock Language-Driven Temporal Activity Localization: {A} Semantic Matching
  Reinforcement Learning Model.
\newblock In \emph{{CVPR}}.

\bibitem[{Wolf et~al.(2019)Wolf, Debut, Sanh, Chaumond, Delangue, Moi, Cistac,
  Rault, Louf, Funtowicz, and Brew}]{DBLP:journals/corr/abs-1910-03771}
Wolf, T.; Debut, L.; Sanh, V.; Chaumond, J.; Delangue, C.; Moi, A.; Cistac, P.;
  Rault, T.; Louf, R.; Funtowicz, M.; and Brew, J. 2019.
\newblock HuggingFace's Transformers: State-of-the-art Natural Language
  Processing.
\newblock \emph{CoRR}, abs/1910.03771.

\bibitem[{Wu et~al.(2020)Wu, Li, Liu, and Lin}]{DBLP:conf/aaai/WuLLL20}
Wu, J.; Li, G.; Liu, S.; and Lin, L. 2020.
\newblock Tree-Structured Policy Based Progressive Reinforcement Learning for
  Temporally Language Grounding in Video.
\newblock In \emph{{AAAI}}.

\bibitem[{Wu et~al.(2018)Wu, Xiong, Yu, and Lin}]{DBLP:conf/cvpr/WuXYL18}
Wu, Z.; Xiong, Y.; Yu, S.~X.; and Lin, D. 2018.
\newblock Unsupervised Feature Learning via Non-Parametric Instance
  Discrimination.
\newblock In \emph{{CVPR}}.

\bibitem[{Xu et~al.(2019)Xu, He, Plummer, Sigal, Sclaroff, and
  Saenko}]{DBLP:conf/aaai/Xu0PSSS19}
Xu, H.; He, K.; Plummer, B.~A.; Sigal, L.; Sclaroff, S.; and Saenko, K. 2019.
\newblock Multilevel Language and Vision Integration for Text-to-Clip
  Retrieval.
\newblock In \emph{{AAAI}}.

\bibitem[{Yu et~al.(2021)Yu, Wang, Hu, Luo, and
  Li}]{DBLP:journals/corr/abs-2106-07166}
Yu, Y.; Wang, X.; Hu, W.; Luo, X.; and Li, C. 2021.
\newblock 2rd Place Solutions in the {HC-STVG} track of Person in Context
  Challenge 2021.
\newblock \emph{CoRR}, abs/2106.07166.

\bibitem[{Yuan et~al.(2019)Yuan, Ma, Wang, Liu, and
  Zhu}]{DBLP:conf/nips/YuanMWL019}
Yuan, Y.; Ma, L.; Wang, J.; Liu, W.; and Zhu, W. 2019.
\newblock Semantic Conditioned Dynamic Modulation for Temporal Sentence
  Grounding in Videos.
\newblock In \emph{NeurIPS}.

\bibitem[{Yuan, Mei, and Zhu(2019)}]{DBLP:conf/aaai/YuanM019}
Yuan, Y.; Mei, T.; and Zhu, W. 2019.
\newblock To Find Where You Talk: Temporal Sentence Localization in Video with
  Attention Based Location Regression.
\newblock In \emph{{AAAI}}.

\bibitem[{Zeng et~al.(2020)Zeng, Xu, Huang, Chen, Tan, and
  Gan}]{DBLP:conf/cvpr/ZengXHCTG20}
Zeng, R.; Xu, H.; Huang, W.; Chen, P.; Tan, M.; and Gan, C. 2020.
\newblock Dense Regression Network for Video Grounding.
\newblock In \emph{{CVPR}}.

\bibitem[{Zhang et~al.(2020{\natexlab{a}})Zhang, Li, Yuan, Xu, Jiang, and
  Shan}]{DBLP:journals/corr/abs-2009-11232}
Zhang, B.; Li, Y.; Yuan, C.; Xu, D.; Jiang, P.; and Shan, Y.
  2020{\natexlab{a}}.
\newblock A Simple Yet Effective Method for Video Temporal Grounding with
  Cross-Modality Attention.
\newblock \emph{CoRR}, abs/2009.11232.

\bibitem[{Zhang et~al.(2019)Zhang, Dai, Wang, Wang, and
  Davis}]{DBLP:conf/cvpr/ZhangDWWD19}
Zhang, D.; Dai, X.; Wang, X.; Wang, Y.; and Davis, L.~S. 2019.
\newblock {MAN:} Moment Alignment Network for Natural Language Moment Retrieval
  via Iterative Graph Adjustment.
\newblock In \emph{{CVPR}}.

\bibitem[{Zhang et~al.(2020{\natexlab{b}})Zhang, Sun, Jing, and
  Zhou}]{DBLP:conf/acl/ZhangSJZ20}
Zhang, H.; Sun, A.; Jing, W.; and Zhou, J.~T. 2020{\natexlab{b}}.
\newblock Span-based Localizing Network for Natural Language Video
  Localization.
\newblock In \emph{{ACL}}.

\bibitem[{Zhang et~al.(2020{\natexlab{c}})Zhang, Peng, Fu, Lu, and
  Luo}]{DBLP:journals/corr/abs-2012-02646}
Zhang, S.; Peng, H.; Fu, J.; Lu, Y.; and Luo, J. 2020{\natexlab{c}}.
\newblock Multi-Scale 2D Temporal Adjacent Networks for Moment Localization
  with Natural Language.
\newblock \emph{CoRR}, abs/2012.02646.

\bibitem[{Zhang et~al.(2020{\natexlab{d}})Zhang, Peng, Fu, and
  Luo}]{DBLP:conf/aaai/ZhangPFL20}
Zhang, S.; Peng, H.; Fu, J.; and Luo, J. 2020{\natexlab{d}}.
\newblock Learning 2D Temporal Adjacent Networks for Moment Localization with
  Natural Language.
\newblock In \emph{{AAAI}}.

\bibitem[{Zhao et~al.(2017)Zhao, Xiong, Wang, Wu, Tang, and
  Lin}]{DBLP:conf/iccv/ZhaoXWWTL17}
Zhao, Y.; Xiong, Y.; Wang, L.; Wu, Z.; Tang, X.; and Lin, D. 2017.
\newblock Temporal Action Detection with Structured Segment Networks.
\newblock In \emph{{ICCV}}.

\bibitem[{Zhao et~al.(2021)Zhao, Zhao, Zhang, and
  Lin}]{DBLP:conf/cvpr/ZhaoZZL21}
Zhao, Y.; Zhao, Z.; Zhang, Z.; and Lin, Z. 2021.
\newblock Cascaded Prediction Network via Segment Tree for Temporal Video
  Grounding.
\newblock In \emph{{CVPR}}.

\end{thebibliography}
\end{document}